\documentclass[10pt,twocolumn,letterpaper]{article}

\usepackage[pagenumbers]{cvpr} 

\usepackage{algorithm}
\usepackage{algorithmic}
\usepackage{amsmath} 
\usepackage{tabularray}
\usepackage{tabularx} 
\usepackage{multirow}
\usepackage{booktabs}
\usepackage{makecell}
\usepackage{array}  
\usepackage{graphicx} 
\usepackage{pgfplots}
\usepackage{pgfplotstable}
\usepackage{tikz} 
\usetikzlibrary{patterns} 
\pgfplotsset{compat=1.18}
\usepackage{xcolor}
\definecolor{myCoral}{HTML}{E99C93}       
\definecolor{myLavenderBlue}{HTML}{9FACD3} 
\definecolor{myPeach}{HTML}{F1DBB9}        
\definecolor{myLilac}{HTML}{D9D1E3}        
\definecolor{myPeriwinkle}{HTML}{CAD4E7}   
\usepackage{subcaption}

\definecolor{cvprblue}{rgb}{0.21,0.49,0.74}
\usepackage[pagebackref,breaklinks,colorlinks,allcolors=cvprblue]{hyperref}

\usepackage{array}
\usepackage{makecell}

\newcolumntype{L}[1]{>{\raggedright\arraybackslash}m{#1}}

\def\paperID{16681} 
\def\confName{CVPR}
\def\confYear{2026}

\vspace{-5mm}

\title{World2VLM: Distilling World Model Imagination into VLMs \\ for Dynamic Spatial Reasoning}

\author{ 
    \textbf{Wanyue Zhang}$^{1,2}$,
    \textbf{Wenxiang Wu}$^{4}$,
    \textbf{Wang Xu}$^{3}$\thanks{Co-corresponding author.},
    \textbf{Jiaxin Luo}$^{4}$, \\
    \textbf{Helu Zhi}$^{4}$, 
    \textbf{Yibin Huang}$^{4}$, 
    \textbf{Shuo Ren}$^{1}$, 
    \textbf{Zitao Liu}$^{4}$, 
    \textbf{Jiajun Zhang}$^{1,2,5}$\footnotemark[1] \\
    $^{1}$Institute of Automation, Chinese Academy of Sciences\\
    $^{2}$School of Artificial Intelligence, University of Chinese Academy of Sciences\\
    $^{3}$Tsinghua University 
    ~$^{4}$Harbin Institute of Technology 
    ~$^{5}$Wuhan AI Research\\
    {\tt\small zhangwanyue2023@ia.ac.cn, xwjim812@gmail.com, shuo.ren@ia.ac.cn, jjzhang@nlpr.ia.ac.cn}
}

\begin{document}

\maketitle

\begin{abstract}
Vision-language models (VLMs) have shown strong performance on static visual understanding, yet they still struggle with dynamic spatial reasoning that requires imagining how scenes evolve under egocentric motion. Recent efforts address this limitation either by scaling spatial supervision with synthetic data or by coupling VLMs with world models at inference time. However, the former often lacks explicit modeling of motion-conditioned state transitions, while the latter incurs substantial computational overhead.
In this work, we propose World2VLM, a training framework that distills spatial imagination from a generative world model into a vision-language model. Given an initial observation and a parameterized camera trajectory, we use a view-consistent world model to synthesize geometrically aligned future views and derive structured supervision for both forward (action-to-outcome) and inverse (outcome-to-action) spatial reasoning.\footnote{The dataset is available at \url{https://huggingface.co/datasets/WanyueZhang/World2VLM}.}
We post-train the VLM with a two-stage recipe on a compact dataset generated by this pipeline and evaluate it on multiple spatial reasoning benchmarks.\footnote{The code is available at \url{https://github.com/WanyueZhang-ai/World2VLM}}
World2VLM delivers consistent improvements over the base model across diverse benchmarks, including SAT-Real, SAT-Synthesized, VSI-Bench, and MindCube. It also outperforms the test-time world-model-coupled methods while eliminating the need for expensive inference-time generation.
Our results suggest that world models can serve not only as inference-time tools, but also as effective training-time teachers, enabling VLMs to internalize spatial imagination in a scalable and efficient manner.
\end{abstract}

\section{Introduction}

\begin{figure}[t!]
\centering
\includegraphics[width=1\linewidth]{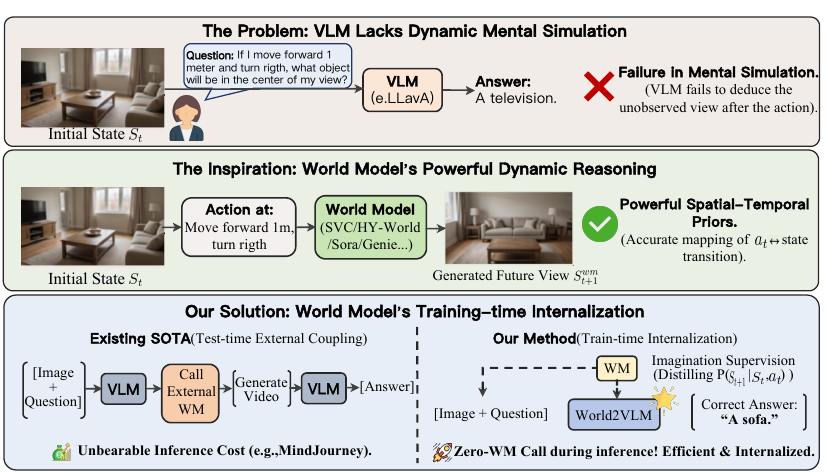}
\caption{Motivation for World2VLM. Existing approaches either rely on static spatial supervision or defer dynamic reasoning to expensive test-time world-model coupling. World2VLM instead uses a generative world model at training time to synthesize action-aligned view transitions and distill dynamic spatial reasoning into the VLM itself.}
\label{fig:dataset}
\end{figure}

Vision-language models (VLMs) have achieved remarkable progress in multimodal understanding, enabling strong performance on tasks such as visual question answering, captioning, and general reasoning over static images. More broadly, spatial understanding is increasingly recognized as a cornerstone capability for modern multimodal models in embodied settings, where it supports perception, navigation, manipulation, and planning \cite{hurst2024gpto,zhu2025internvl3exploringadvancedtraining,li2024llavaov,bai2025qwen25vltechnicalreport,cheng2025embodiedevalevaluatemultimodalllms,du-etal-2024-embspatial,cai2025cookbenchlonghorizonembodiedplanning,li2023manipllmembodiedmultimodallarge,colan2025assessingvaluevisualinput,liu2024world,su2025reactivecognitivebraininspired}. However, despite these advances, current VLMs remain fundamentally limited in \textit{dynamic spatial reasoning}---the ability to infer how a scene evolves under egocentric motion, predict the consequences of actions, or mentally simulate unseen viewpoints. This limitation is particularly evident in tasks involving perspective shifts, multi-step transformations, and action-conditioned reasoning, where models often rely on shallow pattern matching rather than genuine spatial understanding.
Recent camera-centric and ego-centric multi-view studies further suggest that this gap persists even when richer viewpoint information is available, indicating that camera motion itself remains a central reasoning bottleneck \cite{liao2026thinking,gholami2025spatialreasoningvisionlanguagemodels}.

Existing attempts to address this limitation largely follow two directions. The first is to scale spatial supervision through synthetic data, simulation, and broader benchmark construction \cite{chen2024spatialvlm,krishna2024satspatialaptitude,ogezi2025spare,zhang2025flatland,daxberger2025mmspatial,yu2025far,jia2025omnispatialcomprehensivespatialreasoning,yeh2025seeingperspectiveevaluatingmultiview}. While useful, these approaches typically provide static supervision signals and do not explicitly model how observations evolve under actions. The second is to couple VLMs with world models or imagination-based reasoning procedures at inference time \cite{yang2025mindjourney,cao2025spatialdreamer,0001W0ZZLH24,li2025imaginereasoningspacemultimodal,liu2025spatialcotadvancingspatialreasoning,wu2025groundedchainofthoughtmultimodallarge}. Although this can improve dynamic spatial reasoning, it incurs substantial test-time overhead and leaves the VLM itself largely unchanged. This leads to a natural question: 
\begin{quote}
\textit{Can a world model serve as a \textbf{training-time teacher} that distills spatial imagination into a vision-language model?}
\end{quote}
Recent evidence also suggests that world-model priors can be transferred into VLM representations themselves, but existing efforts do not center action-aligned view-transition supervision for dynamic spatial reasoning \cite{zhang2025worldmodelsbenefitvlms}.

In this work, we answer this question with \textbf{World2VLM}, a training framework that treats a generative world model as a \textit{training-time teacher}. Given an anchor observation and a parameterized camera trajectory, we use a view-consistent world model to synthesize geometrically aligned future views and convert them into structured supervision for both inverse spatial reasoning (outcome-to-action) and forward spatial reasoning (action-to-outcome). In this way, spatial imagination is not outsourced to an external module at inference time, but distilled into the parameters of the VLM itself.

World2VLM is organized around three components. First, a \textbf{world-model-guided transition construction pipeline} generates motion-conditioned source-target pairs with explicit action semantics. Second, a \textbf{bidirectional task suite for spatial internalization} converts each transition into complementary supervision for both inverse reasoning (recovering the action behind a view change) and forward reasoning (predicting the consequences of an action). This design is important because dynamic spatial reasoning requires both identifying the latent transition behind observations and mentally propagating that transition to unseen future views. Third, a \textbf{two-stage post-training recipe} combines supervised distillation with task-aware GRPO refinement, enabling the model to internalize dynamic spatial structure and then sharpen its structured spatial outputs.

We evaluate World2VLM on four spatial reasoning benchmarks with substantial dynamic or viewpoint-conditioned components: SAT-Real, SAT-Synthesized, VSI-Bench, and MindCube. Together, they cover both real-image and simulated settings while emphasizing motion-conditioned reasoning, perspective change, and mental simulation. Across all four, World2VLM delivers consistent improvements over the base Qwen2.5-VL model, and the GRPO stage further improves over the SFT checkpoint. Moreover, World2VLM outperforms a MindJourney-style test-time world-model-coupled baseline while requiring only standard VLM inference at deployment. Fine-grained analysis on SAT-Real further shows that the gains are concentrated on subproblems such as egocentric movement, action consequence, and perspective taking, supporting our claim that the method improves motion-conditioned reasoning rather than only static recognition.

Our contributions are summarized as follows:
\begin{itemize}
    \item We propose \textbf{World2VLM}, a training framework that uses a controllable world model as a training-time teacher, distilling motion-conditioned view transitions into VLM parameters for dynamic spatial reasoning.
    \item We introduce a \textbf{bidirectional spatial supervision framework} consisting of a transition-construction distillation pipeline, a forward/inverse task suite, and a two-stage SFT$\rightarrow$GRPO post-training recipe for spatial internalization.
    \item We show through \textbf{comprehensive experiments and analysis} that World2VLM yields consistent gains on SAT-Real, SAT-Synthesized, VSI-Bench, and MindCube, improves the dynamic categories of SAT-Real in particular, and is more effective and robust than test-time world-model coupling.
\end{itemize}

\begin{figure*}[t!]
\centering
\includegraphics[width=\textwidth]{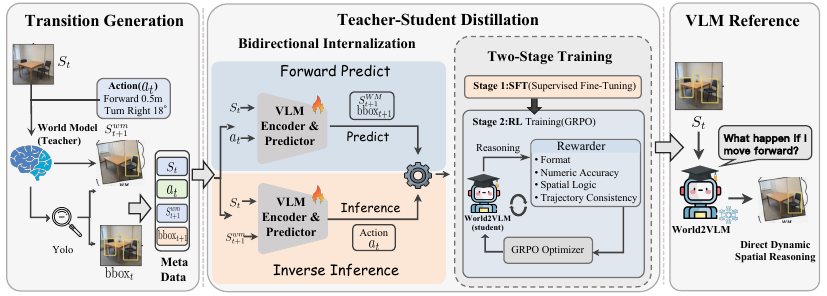}
\caption{Overview of World2VLM. The framework consists of three components. \textbf{(1) World-model-guided transition construction:} starting from an anchor observation and a parameterized egocentric action, a controllable world model synthesizes an action-aligned future view, together with detector-derived object metadata. \textbf{(2) Bidirectional task suite for spatial internalization:} each source-target transition is converted into complementary inverse tasks that recover the underlying motion and forward tasks that predict motion-conditioned consequences. \textbf{(3) Two-stage post-training:} the student VLM is first optimized with supervised fine-tuning on structured spatial supervision and then refined with task-aware GRPO, which rewards valid formatting, numeric accuracy, spatial logic, and trajectory consistency. After training, World2VLM performs dynamic spatial reasoning directly, without querying the world model or generating auxiliary views at inference time.}
\label{fig:method-overview}
\end{figure*}

\section{Related Work}
\subsection{Spatial Reasoning in VLMs}

A growing body of work shows that spatial reasoning remains a core weakness of vision-language models. Existing evaluations span single-view tests, broader multimodal suites, multi-view benchmarks, and dynamic or viewpoint-dependent settings such as SAT, VSI-Bench, MindCube, 3DSRBench, and SITE \cite{kamath-etal-2023-whats,fu2024blink,meng2024mmiu,zhang2025mulset,jia2025omnispatialcomprehensivespatialreasoning,yeh2025seeingperspectiveevaluatingmultiview,krishna2024satspatialaptitude,yang2025thinking,yang2025mmsi,wu2025spatialscore,yin2025mindcube,ma20253dsrbench,wang2025site,ge-etal-2025-mllm}. Camera-centric and ego-centric multi-view analyses reinforce the same conclusion, emphasizing that viewpoint change and camera control remain under-modeled even for strong multimodal systems \cite{liao2026thinking,gholami2025spatialreasoningvisionlanguagemodels}.

On the training side, prior work scales spatial supervision with 3D data, dynamic aptitude tasks, synthetic spatial VQA, and single-view to video curricula \cite{chen2024spatialvlm,krishna2024satspatialaptitude,ogezi2025spare,zhang2025flatland,li2025spatialladder}. More recent post-training approaches activate spatial ability through viewpoint-consistent learning, progressive curricula, explicit 3D representations, structured camera-motion reasoning, visual-intermediate reasoning, and spatially aware architectures \cite{zhan2025actial,li2026spatialladder,ma2025spatialreasoner,wu2026camreasonerreinforcingcameramovement,wu2025reinforcingspatialreasoningvisionlanguage,wu2025spatialmllmboostingmllmcapabilities,he2025spatialormllm,zheng2025multimodal}. Complementary diagnostic and prompting work attributes failures to attention misalignment and weak positional grounding, and explores externalized reasoning through Video-of-Thought, imagination-based prompting, SpatialCoT, grounded CoT, allocentric traces, and explicit 3D exploration \cite{rajabi2024towards,chen2025spatialreasoninghardvlms,qi2025beyond,yang2025cambrian,0001W0ZZLH24,li2025imaginereasoningspacemultimodal,liu2025spatialcotadvancingspatialreasoning,wu2025groundedchainofthoughtmultimodallarge,hua2026unleashingspatialreasoningmultimodal,zhang2026think3dthinkingspacespatial}. Our work instead uses generated action-conditioned transitions as training-time supervision.

\subsection{World Models for Spatial Dynamics}

In this paper, we use generative world models that synthesize future observations conditioned on images, cameras, actions, or video context. Progress in interactive and video world modeling, including Genie, Sora, HunyuanVideo, and camera-controllable generation such as GEN3C, makes this form of supervision increasingly practical \cite{bruce2024genie,openai2024sora,tian2025hunyuanvideosystematicframework,ren2025gen3c,yang2026neoverse}. SVC specializes in camera-conditioned novel views \cite{zhou2025stable}, whereas HY-WorldPlay emphasizes action-conditioned long-horizon dynamics \cite{tian2025hunyuanvideosystematicframework}. Related work further studies action-faithful egocentric prediction, cross-view forecasting, local spatial memories, out-of-sight dynamics, and distillation of world priors into deployable models or policies \cite{bagchi2026walkpaintingsegocentricworld,sharma2026crossviewworldmodels,wang2026anchorweaveworldconsistentvideogeneration,duan2026liveworldsimulatingoutofsightdynamics,zhang2025worldmodelsbenefitvlms,jiang2026wptworldtopolicytransferonline}. We use these models as training-time teachers rather than as inference-time generators.

\subsection{Mental Simulation with World Models}

A closely related line of work studies whether VLMs can reason through imagined views or internal spatial representations. \textbf{MindJourney} is the most direct point of comparison: it couples a VLM with a controllable world model at inference time, allowing the model to iteratively generate imagined views and reason over them without fine-tuning \cite{yang2025mindjourney}. \textbf{SpatialDreamer} follows the same general paradigm, but adds reinforcement learning and geometry-aware optimization on top of world-model-guided imagination \cite{cao2025spatialdreamer}. These methods show that inference-time world-model coupling can improve spatial reasoning, but they also introduce substantial deployment overhead and leave the VLM itself unchanged.

\textbf{MindCube}, by contrast, studies spatial mental modeling from limited views and shows that explicitly constructing intermediate cognitive maps can improve perspective taking and mental simulation \cite{yin2025mindcube}. Together, these works highlight the value of imagined intermediate structure. Our method differs by shifting world-model usage from \emph{test time} to \emph{train time}: rather than invoking an external generator or explicit map at inference time, we use generated future observations as supervision and internalize action-conditioned transitions directly in the VLM.

\section{Method}

\subsection{Overview}

World2VLM treats a controllable world model as a \textit{training-time teacher} for dynamic spatial reasoning. Rather than invoking the world model at inference time, we use it offline to synthesize view transitions under known egocentric motion, convert these transitions into structured reasoning tasks, and train a VLM to internalize the corresponding spatial regularities. The formulation is agnostic to the specific generator used as teacher; in our experiments, we instantiate it with both Stable Virtual Camera and HY-WorldPlay to study how different world-model priors affect downstream spatial supervision.

Figure~\ref{fig:method-overview} summarizes the full teacher-student pipeline. In the data preparation stage, we pair each anchor observation $s_t$ with a sampled egocentric action $a_t$ and use the world model to generate a motion-consistent future view $s_{t+1}^{WM}$. When object-centric supervision is needed, a detector-tracker produces structured metadata that is serialized into the prompt rather than rendered on the image. These transitions are then routed into two complementary supervision paths: a \textit{forward path} that predicts the spatial consequences of an action, and an \textit{inverse path} that infers the action that explains a visual change between views. We first internalize these signals with supervised fine-tuning and then sharpen structured spatial outputs with GRPO. Once training is complete, the world model is no longer needed, and the resulting World2VLM answers dynamic spatial reasoning questions with standard VLM inference alone.

Formally, let $s_t$ denote an anchor observation and $a_t$ denote a parameterized egocentric camera motion. A world model defines a motion-conditioned transition:
\begin{equation}
    s_{t+1}^{WM} \sim P_{WM}(s_{t+1} \mid s_t, a_t),
\end{equation}
where $s_{t+1}^{WM}$ is a synthesized future view aligned with the specified action. Our goal is to train a VLM with parameters $\theta$ to internalize both inverse spatial reasoning and forward spatial consequence prediction:
\begin{equation}
    P_{\theta}(a_t \mid s_t, s_{t+1}) \quad \text{and} \quad P_{\theta}(y \mid s_t, a_t),
\end{equation}
without querying the world model at inference time.

The resulting method has three components: (1) a world-model-guided transition construction pipeline (Sec.~\ref{sec:TransitionConstruction}), (2) a bidirectional task suite for spatial internalization (Sec.~\ref{sec:TaskSuite}), and (3) a two-stage post-training recipe consisting of supervised distillation followed by GRPO refinement (Sec.~\ref{sec:sft}--\ref{sec:grpo}).

\begin{table*}[t]
\centering
\small
\renewcommand{\arraystretch}{1.2}
\begin{tabular}{L{2.6cm}L{3.2cm}L{7.8cm}L{1.6cm}}
\toprule
\textbf{Category} & \textbf{Concrete Task} & \textbf{Task Formulation} & \textbf{Direction} \\
\midrule
Motion reasoning & \makecell[l]{Motion distance\\estimation} & Translation distance from two translationally related views & Inverse \\
  & \makecell[l]{Motion orientation\\estimation} & Turning direction and magnitude from two rotationally related views & Inverse \\
  & \makecell[l]{Multi-step motion\\prediction} & Ordered action sequence for a composed camera trajectory & Inverse \\

\makecell[l]{Object-guided\\motion reasoning} & \makecell[l]{Viewpoint transfor-\\mation inference} & Camera action sequence from cross-view object correspondences & Inverse \\

Action verification & \makecell[l]{Action-sequence\\verification} & Verification of whether a candidate motion matches an image pair & Forward \\

Object grounding & \makecell[l]{Post-action object\\bounding box prediction} & Target bounding box after the described action & Forward \\
  & \makecell[l]{Post-action object\\visibility detection} & Object visibility after the described action & Forward \\

Object consistency & \makecell[l]{Cross-view object\\consistency judgment} & Instance consistency across views under the described action & Forward \\
\bottomrule
\end{tabular}
\caption{Task suite used by World2VLM. Each synthesized transition is converted into one of eight concrete spatial reasoning tasks spanning motion inference, action verification, object grounding, and cross-view consistency. Together, these tasks provide complementary supervision for internalizing motion-conditioned spatial structure.}
\label{tab:method-task-suite}
\end{table*}

\subsection{World-Model-Guided Transition Construction}
\label{sec:TransitionConstruction}

\paragraph{Trajectory-Conditioned View Synthesis.}
Starting from anchor RGB observations, we sample egocentric camera trajectories from a predefined action space that includes translation, rotation, and short multi-step compositions. For each anchor image, the trajectory generator stores both step-wise actions and cumulative motion prefixes. A view-consistent world model then synthesizes future observations along the trajectory:
\begin{equation}
    s_{t+\Delta}^{WM} = \mathcal{G}(s_t, a_t^{(\Delta)}),
\end{equation}
where $a_t^{(\Delta)}$ denotes the cumulative motion up to step $\Delta$. This process yields a controllable sequence of imagined views with explicit action semantics.

\paragraph{Max-Displacement Pairing.}
To make the supervision geometrically informative, we construct training pairs using the anchor frame and the farthest valid synthesized endpoint in the trajectory, rather than relying on short-range neighboring views. This max-displacement pairing amplifies viewpoint change and forces the model to reason over non-trivial spatial transformations instead of local appearance continuity.

\paragraph{Spatial Anchoring.}
For tasks that require object-level reasoning, we augment synthesized transitions with object correspondences obtained from a detector-tracker. Let $B_t$ and $B_{t+\Delta}$ denote the sets of normalized object boxes in the source and target views. Importantly, these boxes are serialized into the text prompt as structured coordinates rather than rendered onto the images. This design allows the model to use object-level anchors while keeping the visual input unchanged. The resulting anchors connect motion to object displacement, visibility change, and cross-view identity, which helps prevent shortcut learning based purely on language priors.

\subsection{Task Suite for Spatial Internalization}
\label{sec:TaskSuite}

World2VLM does not train on a single-question format. Instead, it converts each synthesized transition into a set of complementary spatial reasoning tasks that cover both motion inference and action-conditioned consequence prediction. Table~\ref{tab:method-task-suite} summarizes the task suite used for spatial internalization.

The key design principle is complementarity. Some tasks ask the model to recover the motion that produced a visual transition, whereas others ask it to predict the consequence of a motion on objects, visibility, or identity. Together they force the model to bind action, viewpoint change, and object-level geometry into a unified spatial representation.

It is important to distinguish \textit{task family} from \textit{supervision direction}. The motion-centric family and the object-centric family are grouped by semantic content, whereas the inverse/forward split is grouped by reasoning direction. As a result, the family-based ablations and the direction-based ablations probe different questions and should not be conflated.

\subsection{Stage I: Supervised Distillation with Bidirectional Spatial Supervision}
\label{sec:sft}

We divide the task suite into two supervision streams. The \textbf{inverse stream} trains the model to infer motion from a pair of observations, covering motion distance estimation, rotation angle estimation, multi-step motion inference, and object-guided viewpoint transformation inference. The \textbf{forward stream} trains the model to predict spatial consequences given an initial observation and an action specification, covering action-sequence verification, post-action object localization, post-action visibility judgment, and cross-view object consistency judgment.

Let $\mathcal{D}_{\text{inv}}$ and $\mathcal{D}_{\text{fwd}}$ denote the inverse and forward training sets. We optimize the VLM with a joint supervised objective:
\begin{equation}
    \mathcal{L}_{\text{SFT}} = \lambda_{\text{inv}} \mathcal{L}_{\text{inv}} + \lambda_{\text{fwd}} \mathcal{L}_{\text{fwd}},
\end{equation}
where
\begin{equation}
\begin{aligned}
\mathcal{L}_{\text{inv}} 
&= - \mathbb{E}_{(x,y)\sim \mathcal{D}_{\text{inv}}}\log P_{\theta}(y \mid x), \\
\mathcal{L}_{\text{fwd}} 
&= - \mathbb{E}_{(x,y)\sim \mathcal{D}_{\text{fwd}}}\log P_{\theta}(y \mid x).
\end{aligned}
\end{equation}
Although both terms take the form of language-model supervision, they correspond to fundamentally different spatial queries: recovering latent motion versus forecasting motion-conditioned outcomes. This bidirectional training is the main mechanism by which World2VLM distills world-model imagination into model parameters.

The two directions constrain the same transition from opposite sides. Inverse tasks ask the model to explain \emph{what motion must have happened} given two observations, which encourages sensitivity to viewpoint change and latent camera transformation. Forward tasks ask the model to predict \emph{what should happen next} given an observation and an action, which encourages mental rollout of object displacement, visibility change, and cross-view identity. For example, if a television appears near the right edge of the initial view, a forward task may ask whether it will move toward the center after the camera moves forward and turns right, whereas an inverse task may ask which camera motion best explains why that television becomes centered in the second view. Although both tasks involve the same transition, they require different reasoning operations: one predicts consequences from actions, and the other explains observations by recovering the underlying action. Training on both directions therefore reduces the chance that the model solves either family through shallow shortcuts alone; a useful spatial representation must support both explanation and prediction.

\begin{table*}[t!]
\centering
\small
\setlength{\tabcolsep}{5pt}
\begin{tabular}{llccccc}
\toprule
\textbf{Method} & \textbf{Setting} & \textbf{SAT-Real} & \textbf{SAT-Synth.} & \textbf{VSI-Bench} & \textbf{MindCube} & \textbf{Average} \\
\midrule
Qwen2.5-VL-7B & Base VLM & 44.67 & 39.60 & 33.00 & 29.26 & 36.63 \\
Qwen2.5-VL-7B + WM & MindJourney-style\cite{yang2025mindjourney} & 31.33 (-13.34) & 51.75 (+12.15) & 37.68 (+4.68) & 33.85 (+4.59) & 38.65 (+2.02) \\
\midrule
World2VLM-SFT & SVC as WM & 64.00 (+19.33) & 50.00 (+10.40) & 39.84 (+6.84) & 33.14 (+3.88) & 46.75 (+10.12) \\
World2VLM-GRPO & SVC as WM & \textbf{72.67 (+28.00)} & \underline{59.20 (+19.60)} & \textbf{41.55 (+8.55)} & 34.86 (+5.60) & \underline{52.07 (+15.44)} \\
\midrule
World2VLM-SFT & HY-WorldPlay as WM & 68.66 (+23.99) & 57.20 (+17.60) & 38.63 (+5.63) & \underline{36.57 (+7.31)} & 50.27 (+13.64) \\
World2VLM-GRPO & HY-WorldPlay as WM & \underline{69.33 (+24.66)} & \textbf{65.20 (+25.60)} & \underline{39.07 (+6.07)} & \textbf{36.85 (+7.59)} & \textbf{52.61 (+15.98)} \\
\bottomrule
\end{tabular}
\caption{Main comparison on four dynamic spatial reasoning benchmarks. Results are reported for two teacher world models, SVC and HY-WorldPlay. Across both teachers, World2VLM internalizes world-model guidance at training time, avoids inference-time image generation, and benefits further from task-aware GRPO refinement. The MindJourney-style row is a representative test-time world-model coupling baseline. All scores are reported in percentage units; values in parentheses denote absolute gains or losses in percentage points relative to the base Qwen2.5-VL-7B model. Best results in each column are \textbf{bold}, and second-best results are \underline{underlined}.}
\vspace{-3mm}
\label{tab:main-results}
\end{table*}

\subsection{Stage II: Task-Aware GRPO Refinement}
\label{sec:grpo}

After supervised distillation, we further refine the model with Group Relative Policy Optimization (GRPO). Starting from the SFT policy, we sample multiple responses for the same prompt and assign a unified task-aware reward:
\begin{equation}
    r(\hat{y}, y^\star) = \alpha_{\text{fmt}} r_{\text{fmt}} + \alpha_{\text{sem}} r_{\text{sem}} + \alpha_{\text{num}} r_{\text{num}} + \alpha_{\text{geo}} r_{\text{geo}} + \alpha_{\text{ord}} r_{\text{ord}},
\end{equation}
where inactive terms are set to zero depending on the task. Intuitively, $r_{\text{fmt}}$ enforces valid structured outputs, $r_{\text{sem}}$ captures answer correctness, $r_{\text{num}}$ evaluates distance or angle precision, $r_{\text{geo}}$ measures localization quality for bounding-box prediction, and $r_{\text{ord}}$ evaluates sequence consistency for multi-step motion reasoning. Very long malformed responses are explicitly penalized to discourage verbose but ungrounded generations.
Appendix~\ref{sec:appendix-grpo-reward} gives the exact task-specific reward definitions, coefficient values, parsing rules, and format penalties used in our implementation.
Appendix~\ref{sec:appendix-grpo-analysis} further provides representative SFT$\rightarrow$GRPO case studies that illustrate where the refinement stage improves output quality in practice.

We then optimize the policy with GRPO:
\begin{equation}
    \max_{\theta} \; \mathbb{E}_{\hat{y} \sim \pi_{\theta}(\cdot \mid x)} \left[ A(\hat{y}) \log \pi_{\theta}(\hat{y} \mid x) \right] - \beta \, \mathrm{KL}\!\left(\pi_{\theta} \,\|\, \pi_{\mathrm{ref}}\right),
\end{equation}
where $A(\hat{y})$ is the group-relative advantage induced by the sampled rewards, $\pi_{\mathrm{ref}}$ is the SFT reference policy, and $\beta$ controls KL regularization. This second stage does not replace the distilled supervision; rather, it sharpens the model's decision boundaries on structured spatial outputs, especially for numerically precise actions, ordered action sequences, and object-grounded predictions.

\section{Experiments}

\subsection{Experimental Setup}

\paragraph{Base Model.}
We adopt Qwen2.5-VL-7B-Instruct as the base vision-language model due to its strong visual understanding capabilities and open-source accessibility \cite{bai2025qwen25vltechnicalreport}. All models are initialized from the same checkpoint for fair comparison.

\paragraph{Teacher World Models.}
We instantiate World2VLM with two teacher world models. Our primary teacher is Stable Virtual Camera (SVC), a diffusion-based novel-view synthesis model that generates temporally smooth and geometrically consistent views conditioned on target cameras, making it particularly well suited for motion-conditioned viewpoint transitions \cite{zhou2025stable}. We also study \textbf{HY-WorldPlay}, an action-conditioned interactive world model built on HunyuanVideo, as an alternative teacher. Compared with SVC's direct camera-conditioned view synthesis, HY-WorldPlay provides a complementary video-generation prior with explicit action control and long-horizon consistency \cite{tian2025hunyuanvideosystematicframework}. Using both teachers allows us to test whether World2VLM depends on a specialized view-synthesis engine or can also benefit from a broader action-conditioned world-model prior.

\paragraph{Training Data.}
We construct training data from two complementary RGB sources: real indoor frames from ScanNet \cite{Dai2017ScanNetR3} and simulated indoor scenes from MulSeT \cite{zhang2025mulset}. Starting from an anchor observation and a sampled egocentric action, the teacher world model synthesizes an action-conditioned future view, which we convert into eight task types spanning motion inference, action verification, object localization, visibility judgment, and cross-view consistency. The final supervised fine-tuning set contains about 100K samples, and the GRPO refinement set is a smaller balanced 1K subset built from the same pipeline. Benchmark evaluation files are never mixed into either training stage.

\paragraph{Benchmarks.}
We evaluate on SAT-Real, SAT-Synthesized, VSI-Bench, and MindCube, which together cover both real-image and simulated settings with substantial dynamic or viewpoint-conditioned reasoning. SAT measures dynamic spatial aptitude under egocentric motion; we report both its real-image and synthetic splits \cite{SAT}. VSI-Bench probes configurational, measurement-estimation, and spatiotemporal reasoning from egocentric observations \cite{yang2025thinking}. MindCube emphasizes perspective taking, mental simulation, and action-conditioned reasoning under partial observation \cite{yin2025mindcube}. We follow the official evaluation protocols and exclude benchmark evaluation files from both training stages.

\paragraph{Baselines.}
We compare against four groups of baselines: (1) the base Qwen2.5-VL-7B-Instruct model, (2) a representative test-time world-model coupling baseline that invokes a world model during inference, (3) controlled data baselines that remove or weaken the world-model-based supervision signal, and (4) our two training stages, namely World2VLM-SFT and World2VLM-GRPO. This design lets us compare train-time internalization against inference-time world-model coupling under a backbone-matched setting. 
For the test-time coupling setting, we adapt a MindJourney-style search-and-score pipeline \cite{yang2025mindjourney} to the same frozen Qwen2.5-VL backbone used by our method. The world model generates imagined views under candidate actions, the VLM scores those candidates, and the final answer is produced from the selected trajectory. This baseline captures the standard test-time world-model-coupling paradigm under a backbone-matched setting. This design lets us compare train-time internalization against inference-time world-model coupling under a backbone-matched setting. 

\paragraph{Training Recipe.}
World2VLM is trained in two stages: supervised fine-tuning followed by task-aware GRPO refinement. We use LoRA-style parameter-efficient adaptation \cite{hu2022lora,xu2023parameter} rather than full-parameter tuning, and the world model is used only during offline data generation. At inference time, the resulting model performs standard VLM decoding without querying the world model.
We take care to separate training supervision from benchmark evaluation. The world-model-generated SFT and GRPO data are constructed from source RGB images and trajectory metadata, rather than from benchmark question-answer files. During post-training, benchmark evaluation instances are not used for optimization, model selection, or prompt-template tuning. 

Detailed trajectory construction, task templates, filtering rules, reward definitions, hyperparameters, and baseline configurations are deferred to Appendix~\ref{sec:appendix-data-construction}, \ref{sec:appendix-task-templates}, \ref{sec:appendix-grpo-reward}, \ref{sec:appendix-implementation-details}, and \ref{sec:appendix-baseline-details} for readability.

\begin{figure}[t]
\centering
\includegraphics[width=\linewidth]{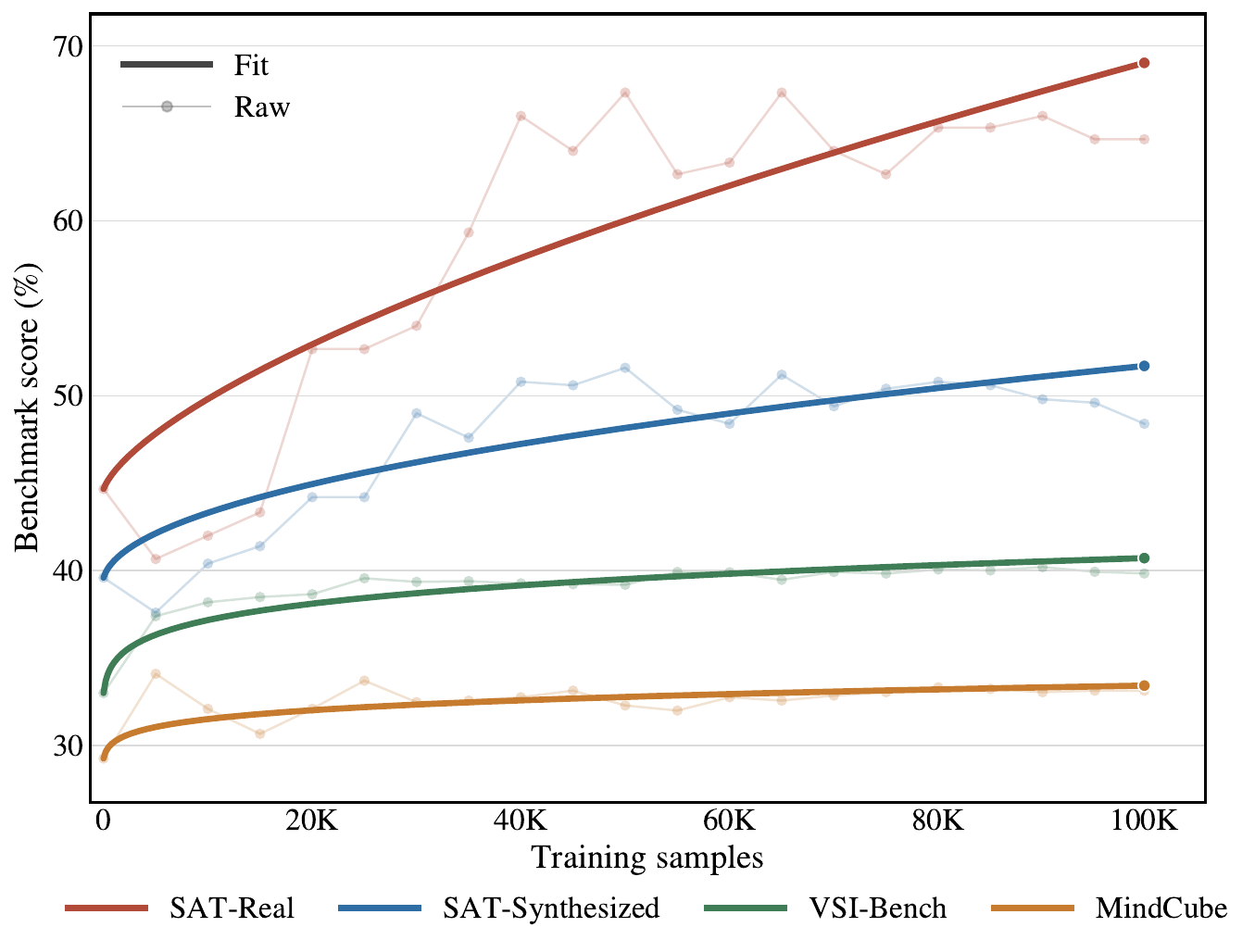}
\caption{Absolute benchmark scores of World2VLM-SFT on four dynamic spatial reasoning benchmarks as a function of the amount of world-model-generated training data (0--100K samples). Faint lines and markers denote raw checkpoint evaluations, while solid curves show power-curve fits that smooth short-term fluctuations. Performance improves rapidly in the low-data regime and then gradually saturates, with the clearest gains on SAT-Real and SAT-Synthesized.}
\vspace{-3mm}
\label{fig:scaling-sft}
\end{figure}

\begin{table*}[t]
\centering
\footnotesize
\setlength{\tabcolsep}{4pt}
\begin{tabular}{llcccccc}
\toprule
\textbf{Method} & \textbf{Setting} & \multicolumn{6}{c}{\textbf{SAT-Real}} \\
\cmidrule(lr){3-8}
 &  & \textbf{All Acc.} & \textbf{Ego Move.} & \textbf{Obj. Move.} & \textbf{Goal Aim} & \textbf{Action Cons.} & \textbf{Perspective} \\
\midrule
Qwen2.5-VL-7B & Base VLM & 44.67 & 52.17 & 21.74 & 50.00 & 45.95 & 48.48 \\
Qwen2.5-VL-7B + WM & Test-time baseline & 31.33 (-13.34) & 35.00 (-17.17) & 45.00 (+23.26) & 16.67 (-33.33) & 34.29 (-11.66) & 34.38 (-14.10) \\
\midrule
World2VLM-SFT & SVC as WM & 64.00 (+19.33) & 73.91 (+21.74) & 43.48 (+21.74) & 79.41 (+29.41) & 75.68 (+29.73) & 42.42 (-6.06) \\
World2VLM-GRPO & SVC as WM & \textbf{72.67 (+28.00)} & \textbf{100.00 (+47.83)} & \textbf{65.22 (+43.48)} & \textbf{79.41 (+29.41)} & \textbf{67.57 (+21.62)} & \textbf{57.57 (+9.09)} \\
\bottomrule
\end{tabular}
\caption{Task-level breakdown on SAT-Real. In addition to overall SAT-Real accuracy, we report five dynamic categories: egocentric movement, object movement, goal aim, action consequence, and perspective taking. All scores are reported in percentage units; values in parentheses denote absolute gains or losses in percentage points relative to the base Qwen2.5-VL-7B model. Best results in each column are \textbf{bold}.}
\label{tab:sat-real-breakdown}
\end{table*}

\begin{table*}[t]
\centering
\small
\setlength{\tabcolsep}{5pt}
\textbf{(a) Distillation Direction Ablation}

\begin{tabular}{lccccc}
\toprule
\textbf{Variant} & \textbf{SAT-R} & \textbf{SAT-S} & \textbf{VSI} & \textbf{MindCube} & \textbf{Avg.} \\
\midrule
Base Model & 44.67 & 39.60 & 33.00 & 29.26 & 36.63 \\
\midrule
Forward Only & 52.00 (+7.33) & \textbf{50.40 (+10.80)} & 38.03 (+5.03) & \textbf{33.33 (+4.07)} & 43.44 (+6.81) \\
Inverse Only & 56.67 (+12.00) & 38.60 (-1.00) & 39.51 (+6.51) & 31.05 (+1.79) & 41.46 (+4.83) \\
Bidirectional (Full SFT) & \textbf{64.00 (+19.33)} & 50.00 (+10.40) & \textbf{39.84 (+6.84)} & 33.14 (+3.88) & \textbf{46.75 (+10.12)} \\
\bottomrule
\end{tabular}

\vspace{0.6em}

\textbf{(b) Training Source Composition Ablation}

\begin{tabular}{lccccc}
\toprule
\textbf{Variant} & \textbf{SAT-R} & \textbf{SAT-S} & \textbf{VSI} & \textbf{MindCube} & \textbf{Avg.} \\
\midrule
Base Model & 44.67 & 39.60 & 33.00 & 29.26 & 36.63 \\
\midrule
Real-Scene Only & 53.33 (+8.66) & 43.00 (+3.40) & 38.95 (+5.95) & 33.52 (+4.26) & 42.20 (+5.57) \\
Simulated-Scene Only & 51.33 (+6.66) & 47.80 (+8.20) & 38.96 (+5.96) & 32.57 (+3.31) & 42.67 (+6.04) \\
Mixed-Source (Full SFT) & \textbf{64.00 (+19.33)} & \textbf{50.00 (+10.40)} & \textbf{39.84 (+6.84)} & 33.14 (+3.88) & \textbf{46.75 (+10.12)} \\
\bottomrule
\end{tabular}
\caption{Ablation of SFT design choices, compared against the base Qwen2.5-VL-7B model. The (a) subtable isolates the effect of distillation direction, while the (b) subtable isolates the effect of training-source composition. All scores are reported in percentage units; values in parentheses denote absolute gains or losses in percentage points relative to the base model. Best results in each column are \textbf{bold}.}
\label{tab:forward-inverse-ablation}
\end{table*}

\subsection{Overall Results}

Table~\ref{tab:main-results} compares World2VLM with the base Qwen2.5-VL-7B model and a representative MindJourney-style test-time world-model-coupled baseline.

The main result is that \emph{training-time internalization is more effective than a representative test-time coupling baseline} across both teacher choices. With SVC as the teacher, World2VLM-GRPO performs best on SAT-Real and VSI-Bench. With HY-WorldPlay, it performs best on SAT-Synthesized, MindCube, and the overall average. This pattern suggests that the framework transfers across distinct world-model priors rather than depending on a single generator family.

The teacher comparison is also informative. SVC is strongest on the benchmarks that depend more directly on camera-conditioned viewpoint consistency, especially SAT-Real and VSI-Bench. HY-WorldPlay, by contrast, is stronger on SAT-Synthesized and MindCube, suggesting that its action-conditioned temporal prior is particularly useful for synthetic dynamics and mental simulation. This difference changes where the gains are largest, but not the overall conclusion that train-time internalization remains effective across teacher families.

The MindJourney-style baseline is still meaningful: it improves over the base model on SAT-Synthesized, VSI-Bench, MindCube, and average score (+2.02). However, all four train-time variants achieve larger average gains while requiring only standard VLM inference at deployment, yielding a better accuracy-versus-deployment-cost trade-off than invoking the generator during inference.

Combined with the ablations in Sec.~\ref{sec:ablation_studies}, these gains suggest that the benefit does not come merely from adding more synthetic supervision, but from training the model to support both explanation of observed transitions and prediction of unseen ones.

Figure~\ref{fig:scaling-sft} shows how performance changes as we increase the amount of world-model-generated SFT data from 0 to 100K samples. Although the raw checkpoint evaluations are mildly non-monotonic, the power-fit curves reveal a clear underlying scaling trend: benchmark scores improve quickly in the low-data regime and then exhibit diminishing returns as more data are added. The effect is strongest on SAT-Real and SAT-Synthesized, where success depends more directly on motion-conditioned viewpoint reasoning, while VSI-Bench and MindCube show smaller but still consistent gains. This pattern suggests that world-model-generated transition supervision provides scalable benefits for dynamic spatial reasoning, especially on tasks that require mentally propagating scene changes under egocentric motion.

\subsection{SAT-Real Task Breakdown}

SAT-Real is particularly informative because it decomposes dynamic spatial reasoning under egocentric motion into several task categories. We therefore analyze five of its representative dynamic categories separately: \emph{egocentric movement}, which requires updating the scene under camera motion; \emph{object movement}, which tracks whether and how an object moves across views; \emph{goal aim}, which asks whether a target becomes reachable or visible under an action; \emph{action consequence}, which predicts how a spatial relation changes after an action; and \emph{perspective taking}, which requires reasoning from another agent's viewpoint. This task-level taxonomy is introduced here because its primary purpose is analytical: it clarifies which motion-sensitive abilities drive the overall SAT-Real improvement.

Table~\ref{tab:sat-real-breakdown} shows that the SAT-Real gain is broad but not perfectly uniform across the benchmark's dynamic categories. Even before RL refinement, World2VLM-SFT already strengthens four of the five categories, especially those that require explicit reasoning about camera motion and action-conditioned scene change, while perspective taking remains slightly below the base model. This pattern is well aligned with the supervision used in World2VLM: the model is trained not only to recover motion from a pair of views, but also to predict what will happen after an action is taken.

The GRPO stage broadens this improvement further and converts the SAT-Real gain into a genuinely task-level effect rather than a single-category spike. In particular, the refinement stage is what lifts perspective taking from slightly below the base model to clearly above it, suggesting that this category benefits from better-calibrated structured reasoning in addition to transition-based supervision. At the same time, the action-consequence score after GRPO is lower than the SFT peak, although it remains substantially above the base model. Overall, the strongest gains still appear on egocentric movement, goal-related reasoning, object movement, and action consequence prediction, which are exactly the settings where a model must mentally propagate scene changes rather than recognize a static relation.

This breakdown also suggests that the gains are not uniformly distributed across all forms of dynamic reasoning. Categories such as egocentric movement and goal aim benefit strongly from transition supervision alone, whereas perspective taking appears to need both transition supervision and the additional calibration provided by RL refinement. In contrast, action consequence already improves substantially after SFT, so the smaller GRPO gain there is consistent with a setting that is easier to internalize during supervised distillation than during reward-based sharpening.

By contrast, the test-time world-model baseline remains category-dependent across the same tasks: it helps object movement, but degrades several other dynamic tasks and lowers the overall SAT-Real score. This breakdown reinforces the main conclusion of the paper: World2VLM improves SAT-Real not by exploiting static shortcuts, but by more reliably internalizing dynamic spatial reasoning.

\subsection{Ablation Studies}
\label{sec:ablation_studies}

\subsubsection{Effect of Bidirectional Distillation}

We first ask whether the two distillation directions are complementary or whether one direction alone is sufficient. The left subtable in Table~\ref{tab:forward-inverse-ablation} shows that they exhibit clearly different strengths. Forward-only distillation is relatively stronger on SAT-Synthesized and MindCube, where the model is directly asked to predict the consequences of an action or mentally simulate unseen outcomes. Inverse-only distillation is stronger on SAT-Real and VSI-Bench, where success more often depends on recovering the egocentric motion that explains a viewpoint change.

Their combination yields the strongest overall result, indicating that the two directions are not redundant. Intuitively, inverse tasks teach the model to decode the latent motion behind a visual transition, while forward tasks teach it to roll that motion forward and anticipate its consequences on visibility, object position, and cross-view consistency. For example, if a television moves from the right edge of the first view to the center of the second, inverse distillation asks which action best explains that change, whereas forward distillation asks whether a specified action would produce it. Bidirectionality is therefore not merely a task-diversification trick; it imposes complementary constraints on the same action-conditioned transition and encourages a spatial representation that supports both explanation and prediction.

\subsubsection{Effect of Training Source Composition}

We next examine whether World2VLM mainly benefits from one source domain or from combining both. The right subtable in Table~\ref{tab:forward-inverse-ablation} shows that real-scene-only training improves SAT-Real more strongly, while simulated-scene-only training is stronger on SAT-Synthesized. However, both source-restricted variants fall short of the mixed-source setting in average score.

This pattern suggests that the two domains provide complementary supervision. Real scenes expose the model to clutter, occlusion, and appearance variation, whereas simulated scenes provide cleaner and more controllable viewpoint-conditioned transitions. Mixing them appears to improve the robustness of the learned spatial transition prior across both real and synthetic evaluation settings.

\section{Conclusion}

In this work, we proposed \textbf{World2VLM}, a training framework that uses world-model-generated view transitions as supervision for dynamic spatial reasoning. By combining inverse and forward spatial learning in a two-stage SFT$\rightarrow$GRPO recipe, World2VLM internalizes how scenes change under egocentric actions instead of relying on inference-time imagination.

The core idea is to treat the world model as a \textit{training-time teacher}. Motion-conditioned future views provide a practical source of supervision for learning both motion inference and action consequence prediction, while the subsequent GRPO stage sharpens structured spatial outputs after this transition knowledge has been distilled into the model parameters.

Empirically, World2VLM improves the base Qwen2.5-VL model on SAT-Real, SAT-Synthesized, VSI-Bench, and MindCube, with especially clear gains on motion-conditioned reasoning. The framework also transfers across two different teacher families, SVC and HY-WorldPlay, and outperforms the MindJourney-style test-time coupling baseline while using standard VLM inference at deployment. These results support our central claim that treating world models as training-time teachers is an effective and practical way to equip VLMs with dynamic spatial reasoning.

\newpage

{
    \small
    \bibliographystyle{ieeenat_fullname}
    \bibliography{main}

\begin{thebibliography}{69}
\providecommand{\natexlab}[1]{#1}
\providecommand{\url}[1]{\texttt{#1}}
\expandafter\ifx\csname urlstyle\endcsname\relax
  \providecommand{\doi}[1]{doi: #1}\else
  \providecommand{\doi}{doi: \begingroup \urlstyle{rm}\Url}\fi

\bibitem[Bagchi et~al.(2026)Bagchi, Bao, Bharadhwaj, Wang, Tokmakov, and Hebert]{bagchi2026walkpaintingsegocentricworld}
Anurag Bagchi, Zhipeng Bao, Homanga Bharadhwaj, Yu-Xiong Wang, Pavel Tokmakov, and Martial Hebert.
\newblock Walk through paintings: Egocentric world models from internet priors, 2026.

\bibitem[Bai et~al.(2025)Bai, Chen, Liu, Wang, Ge, Song, Dang, Wang, Wang, Tang, Zhong, Zhu, Yang, Li, Wan, Wang, Ding, Fu, Xu, Ye, Zhang, Xie, Cheng, Zhang, Yang, Xu, and Lin]{bai2025qwen25vltechnicalreport}
Shuai Bai, Keqin Chen, Xuejing Liu, Jialin Wang, Wenbin Ge, Sibo Song, Kai Dang, Peng Wang, Shijie Wang, Jun Tang, Humen Zhong, Yuanzhi Zhu, Mingkun Yang, Zhaohai Li, Jianqiang Wan, Pengfei Wang, Wei Ding, Zheren Fu, Yiheng Xu, Jiabo Ye, Xi Zhang, Tianbao Xie, Zesen Cheng, Hang Zhang, Zhibo Yang, Haiyang Xu, and Junyang Lin.
\newblock Qwen2.5-vl technical report, 2025.

\bibitem[Brooks et~al.(2024)Brooks, Peebles, Holmes, DePue, Guo, Jing, Schnurr, Taylor, Luhman, Luhman, et~al.]{openai2024sora}
Tim Brooks, Bill Peebles, Connor Holmes, Will DePue, Yufei Guo, Leo Jing, David Schnurr, Joe Taylor, Troy Luhman, Eric Luhman, et~al.
\newblock Video generation models as world simulators.
\newblock \emph{OpenAI Blog}, 1\penalty0 (8):\penalty0 1, 2024.

\bibitem[Bruce et~al.(2024)Bruce, Dennis, Edwards, Parker-Holder, Shi, Hughes, Lai, Mavalankar, Steigerwald, Apps, et~al.]{bruce2024genie}
Jake Bruce, Michael~D Dennis, Ashley Edwards, Jack Parker-Holder, Yuge Shi, Edward Hughes, Matthew Lai, Aditi Mavalankar, Richie Steigerwald, Chris Apps, et~al.
\newblock Genie: Generative interactive environments.
\newblock In \emph{International Conference on Machine Learning}, 2024.

\bibitem[Cai et~al.(2025)Cai, Chen, An, Zhang, Wang, Xu, Zhang, and Liu]{cai2025cookbenchlonghorizonembodiedplanning}
Muzhen Cai, Xiubo Chen, Yining An, Jiaxin Zhang, Xuesong Wang, Wang Xu, Weinan Zhang, and Ting Liu.
\newblock Cookbench: A long-horizon embodied planning benchmark for complex cooking scenarios, 2025.

\bibitem[Cao et~al.(2025)Cao, Li, Liu, Reid, and Liang]{cao2025spatialdreamer}
Meng Cao, Xingyu Li, Xue Liu, Ian Reid, and Xiaodan Liang.
\newblock Spatialdreamer: Incentivizing spatial reasoning via active mental imagery.
\newblock \emph{arXiv preprint arXiv:2512.07733}, 2025.

\bibitem[Chen et~al.(2024)Chen, Xu, Kirmani, Ichter, Driess, Florence, Sadigh, Guibas, and Xia]{chen2024spatialvlm}
Boyuan Chen, Zhuo Xu, Sean Kirmani, Brian Ichter, Danny Driess, Pete Florence, Dorsa Sadigh, Leonidas~J. Guibas, and Fei Xia.
\newblock Spatialvlm: Endowing vision-language models with spatial reasoning capabilities.
\newblock \emph{IEEE/CVF Conference on Computer Vision and Pattern Recognition (CVPR)}, pages 14455--14465, 2024.

\bibitem[Chen et~al.(2025)Chen, Zhu, Zhou, Zhang, Gao, Niebles, Geva, He, Wu, and Li]{chen2025spatialreasoninghardvlms}
Shiqi Chen, Tongyao Zhu, Ruochen Zhou, Jinghan Zhang, Siyang Gao, Juan~Carlos Niebles, Mor Geva, Junxian He, Jiajun Wu, and Manling Li.
\newblock Why is spatial reasoning hard for vlms? an attention mechanism perspective on focus areas.
\newblock In \emph{Proceedings of the International Conference on Machine Learning}, 2025.

\bibitem[Cheng et~al.(2025)Cheng, Tu, Li, Dai, Hu, Hu, Li, Shi, Yu, Chen, Shi, and Sun]{cheng2025embodiedevalevaluatemultimodalllms}
Zhili Cheng, Yuge Tu, Ran Li, Shiqi Dai, Jinyi Hu, Shengding Hu, Jiahao Li, Yang Shi, Tianyu Yu, Weize Chen, Lei Shi, and Maosong Sun.
\newblock Embodiedeval: Evaluate multimodal llms as embodied agents.
\newblock In \emph{The Annual Meeting of the Association for Computational Linguistics}, 2025.

\bibitem[Colan et~al.(2025)Colan, Davila, and Hasegawa]{colan2025assessingvaluevisualinput}
Jacinto Colan, Ana Davila, and Yasuhisa Hasegawa.
\newblock Assessing the value of visual input: A benchmark of multimodal large language models for robotic path planning, 2025.

\bibitem[Dai et~al.(2017)Dai, Chang, Savva, Halber, Funkhouser, and Nießner]{Dai2017ScanNetR3}
Angela Dai, Angel~X. Chang, M. Savva, Maciej Halber, T. Funkhouser, and M. Nießner.
\newblock Scannet: Richly-annotated 3d reconstructions of indoor scenes.
\newblock \emph{IEEE Conference on Computer Vision and Pattern Recognition (CVPR)}, pages 2432--2443, 2017.

\bibitem[Daxberger et~al.(2025)Daxberger, Wenzel, Griffiths, Gang, Lazarow, Kohavi, Kang, Eichner, Yang, Dehghan, and Grasch]{daxberger2025mmspatial}
Erik Daxberger, Nina Wenzel, David Griffiths, Haiming Gang, Justin Lazarow, Gefen Kohavi, Kai Kang, Marcin Eichner, Yinfei Yang, Afshin Dehghan, and Peter Grasch.
\newblock Mm-spatial: Exploring 3d spatial understanding in multimodal llms.
\newblock \emph{arXiv preprint arXiv:2503.13111}, 2025.

\bibitem[Du et~al.(2024)Du, Wu, Li, Huang, and Wei]{du-etal-2024-embspatial}
Mengfei Du, Binhao Wu, Zejun Li, Xuanjing Huang, and Zhongyu Wei.
\newblock {E}mb{S}patial-bench: Benchmarking spatial understanding for embodied tasks with large vision-language models.
\newblock In \emph{Proceedings of the Annual Meeting of the Association for Computational Linguistics}, pages 346--355, 2024.

\bibitem[Duan et~al.(2026)Duan, Xia, Zhang, Zhang, Zhou, Gou, He, Chen, Zhang, and Liu]{duan2026liveworldsimulatingoutofsightdynamics}
Zicheng Duan, Jiatong Xia, Zeyu Zhang, Wenbo Zhang, Gengze Zhou, Chenhui Gou, Yefei He, Feng Chen, Xinyu Zhang, and Lingqiao Liu.
\newblock Liveworld: Simulating out-of-sight dynamics in generative video world models, 2026.

\bibitem[Fei et~al.(2024)Fei, Wu, Ji, Zhang, Zhang, Lee, and Hsu]{0001W0ZZLH24}
Hao Fei, Shengqiong Wu, Wei Ji, Hanwang Zhang, Meishan Zhang, Mong-Li Lee, and Wynne Hsu.
\newblock Video-of-thought: Step-by-step video reasoning from perception to cognition.
\newblock In \emph{Proceedings of the International Conference on Machine Learning}, 2024.

\bibitem[Fu et~al.(2024)Fu, Hu, Li, Feng, Wang, Lin, Roth, Smith, Ma, and Krishna]{fu2024blink}
Xingyu Fu, Yushi Hu, Bangzheng Li, Yu Feng, Haoyu Wang, Xudong Lin, Dan Roth, Noah~A Smith, Wei-Chiu Ma, and Ranjay Krishna.
\newblock Blink: Multimodal large language models can see but not perceive.
\newblock In \emph{European Conference on Computer Vision}, pages 148--166. Springer, 2024.

\bibitem[Ge et~al.(2025)Ge, Chen, Chen, Chen, Chen, Chen, Xie, Yan, Zhu, Lin, Song, Wang, Gao, Zhiyi, Li, Wan, and Wang]{ge-etal-2025-mllm}
Wentao Ge, Shunian Chen, Hardy Chen, Nuo Chen, Junying Chen, Zhihong Chen, Wenya Xie, Shuo Yan, Chenghao Zhu, Ziyue Lin, Dingjie Song, Xidong Wang, Anningzhe Gao, Zhang Zhiyi, Jianquan Li, Xiang Wan, and Benyou Wang.
\newblock {MLLM}-bench: Evaluating multimodal {LLM}s with per-sample criteria.
\newblock In \emph{Proceedings of the Conference of the Nations of the Americas Chapter of the Association for Computational Linguistics}, pages 4951--4974, 2025.

\bibitem[Gholami et~al.(2025)Gholami, Rezaei, Weimin, Mao, Zhou, Zhang, and Akbari]{gholami2025spatialreasoningvisionlanguagemodels}
Mohsen Gholami, Ahmad Rezaei, Zhou Weimin, Sitong Mao, Shunbo Zhou, Yong Zhang, and Mohammad Akbari.
\newblock Spatial reasoning with vision-language models in ego-centric multi-view scenes, 2025.

\bibitem[He et~al.(2025)He, Zhang, Zhang, Zhao, and Peng]{he2025spatialormllm}
Peiqi He, Zhenhao Zhang, Yixiang Zhang, Xiongjun Zhao, and Shaoliang Peng.
\newblock Spatial-ormllm: Improve spatial relation understanding in the operating room with multimodal large language model.
\newblock \emph{arXiv preprint arXiv:2508.08199}, 2025.

\bibitem[Hu et~al.(2022)Hu, Shen, Wallis, Allen-Zhu, Li, Wang, Wang, Chen, et~al.]{hu2022lora}
Edward~J Hu, Yelong Shen, Phillip Wallis, Zeyuan Allen-Zhu, Yuanzhi Li, Shean Wang, Lu Wang, Weizhu Chen, et~al.
\newblock Lora: Low-rank adaptation of large language models.
\newblock \emph{International Conference on Learning Representations}, 1\penalty0 (2):\penalty0 3, 2022.

\bibitem[Hua et~al.(2026)Hua, Yin, Wu, Wang, Huang, and Liu]{hua2026unleashingspatialreasoningmultimodal}
Jiacheng Hua, Yishu Yin, Yuhang Wu, Tai Wang, Yifei Huang, and Miao Liu.
\newblock Unleashing spatial reasoning in multimodal large language models via textual representation guided reasoning, 2026.

\bibitem[Hurst et~al.(2024)Hurst, Lerer, Goucher, Perelman, Ramesh, Clark, Ostrow, Welihinda, Hayes, Radford, et~al.]{hurst2024gpto}
Aaron Hurst, Adam Lerer, Adam~P Goucher, Adam Perelman, Aditya Ramesh, Aidan Clark, AJ Ostrow, Akila Welihinda, Alan Hayes, Alec Radford, et~al.
\newblock Gpt-4o system card.
\newblock \emph{arXiv preprint arXiv:2410.21276}, 2024.

\bibitem[Jia et~al.(2025)Jia, Qi, Zhang, Zhang, Yu, He, Wang, and Yi]{jia2025omnispatialcomprehensivespatialreasoning}
Mengdi Jia, Zekun Qi, Shaochen Zhang, Wenyao Zhang, Xinqiang Yu, Jiawei He, He Wang, and Li Yi.
\newblock Omnispatial: Towards comprehensive spatial reasoning benchmark for vision language models, 2025.

\bibitem[Jiang et~al.(2026)Jiang, Luo, Liu, Huang, Zhu, Qu, Chen, Liu, and Yan]{jiang2026wptworldtopolicytransferonline}
Guangfeng Jiang, Yueru Luo, Jun Liu, Yi Huang, Yiyao Zhu, Zhan Qu, Dave~Zhenyu Chen, Bingbing Liu, and Xu Yan.
\newblock Wpt: World-to-policy transfer via online world model distillation, 2026.

\bibitem[Kamath et~al.(2023)Kamath, Hessel, and Chang]{kamath-etal-2023-whats}
Amita Kamath, Jack Hessel, and Kai-Wei Chang.
\newblock What{'}s ``up'' with vision-language models? investigating their struggle with spatial reasoning.
\newblock In \emph{Proceedings of the Conference on Empirical Methods in Natural Language Processing}, pages 9161--9175, 2023.

\bibitem[Krishna et~al.(2024)Krishna, Duan, Ehsani, Saenko, Kembhavi, Bashkirova, Plummer, Hendrix, Zeng, Tan, and Ray]{krishna2024satspatialaptitude}
Ranjay Krishna, Jiafei Duan, Kiana Ehsani, Kate Saenko, Aniruddha Kembhavi, Dina Bashkirova, Bryan~A. Plummer, Rose Hendrix, Kuo-Hao Zeng, Reuben Tan, and Arijit Ray.
\newblock Sat: Spatial aptitude training for multimodal language models, 2024.

\bibitem[Li et~al.(2024)Li, Zhang, Guo, Zhang, Li, Zhang, Zhang, Li, Liu, and Li]{li2024llavaov}
Bo Li, Yuanhan Zhang, Dong Guo, Renrui Zhang, Feng Li, Hao Zhang, Kaichen Zhang, Yanwei Li, Ziwei Liu, and Chunyuan Li.
\newblock Llava-onevision: Easy visual task transfer.
\newblock \emph{arXiv preprint arXiv:2408.03326}, 2024.

\bibitem[Li et~al.(2025{\natexlab{a}})Li, Wu, Zhang, Xia, Mao, Dong, Vulić, and Wei]{li2025imaginereasoningspacemultimodal}
Chengzu Li, Wenshan Wu, Huanyu Zhang, Yan Xia, Shaoguang Mao, Li Dong, Ivan Vulić, and Furu Wei.
\newblock Imagine while reasoning in space: Multimodal visualization-of-thought.
\newblock In \emph{Proceedings of the International Conference on Machine Learning}, 2025{\natexlab{a}}.

\bibitem[Li et~al.(2025{\natexlab{b}})Li, Li, Wang, Yan, Wu, Zhang, Shen, Lu, Xiao, and Zhuang]{li2025spatialladder}
Hongxing Li, Dingming Li, Zixuan Wang, Yuchen Yan, Hang Wu, Wenqi Zhang, Yongliang Shen, Weiming Lu, Jun Xiao, and Yueting Zhuang.
\newblock Spatialladder: Progressive training for spatial reasoning in vision-language models.
\newblock \emph{arXiv preprint arXiv:2510.08531}, 2025{\natexlab{b}}.

\bibitem[Li et~al.(2026)Li, Li, Wang, Yan, Wu, Zhang, Shen, Lu, Xiao, and Zhuang]{li2026spatialladder}
Hongxing Li, Dingming Li, Zixuan Wang, Yuchen Yan, Hang Wu, Wenqi Zhang, Yongliang Shen, Weiming Lu, Jun Xiao, and Yueting Zhuang.
\newblock Spatialladder: Progressive training for spatial reasoning in vision-language models.
\newblock In \emph{The International Conference on Learning Representations}, 2026.

\bibitem[Li et~al.(2023)Li, Zhang, Geng, Geng, Long, Shen, Zhang, Liu, and Dong]{li2023manipllmembodiedmultimodallarge}
Xiaoqi Li, Mingxu Zhang, Yiran Geng, Haoran Geng, Yuxing Long, Yan Shen, Renrui Zhang, Jiaming Liu, and Hao Dong.
\newblock Manipllm: Embodied multimodal large language model for object-centric robotic manipulation, 2023.

\bibitem[Liao et~al.(2026)Liao, Wu, Wu, Jin, Wang, Wang, Wang, Li, and Loy]{liao2026thinking}
Kang Liao, Size Wu, Zhonghua Wu, Linyi Jin, Chao Wang, Yikai Wang, Fei Wang, Wei Li, and Chen~Change Loy.
\newblock Thinking with camera: A unified multimodal model for camera-centric understanding and generation.
\newblock In \emph{The International Conference on Learning Representations}, 2026.

\bibitem[Liu et~al.(2024)Liu, Yan, Zaharia, and Abbeel]{liu2024world}
Hao Liu, Wilson Yan, Matei Zaharia, and Pieter Abbeel.
\newblock World model on million-length video and language with ringattention.
\newblock \emph{arXiv preprint arXiv:2402.08268}, 2024.

\bibitem[Liu et~al.(2025)Liu, Chi, Wu, Zhang, Hu, Zhang, Zhang, Wu, Cao, Huang, Huang, Tian, Qiu, Quan, Hao, and Zhuang]{liu2025spatialcotadvancingspatialreasoning}
Yuecheng Liu, Dafeng Chi, Shiguang Wu, Zhanguang Zhang, Yaochen Hu, Lingfeng Zhang, Yingxue Zhang, Shuang Wu, Tongtong Cao, Guowei Huang, Helong Huang, Guangjian Tian, Weichao Qiu, Xingyue Quan, Jianye Hao, and Yuzheng Zhuang.
\newblock Spatialcot: Advancing spatial reasoning through coordinate alignment and chain-of-thought for embodied task planning, 2025.

\bibitem[Ma et~al.(2025{\natexlab{a}})Ma, Chen, Zhang, Chou, Chen, de~Melo, and Yuille]{ma20253dsrbench}
Wufei Ma, Haoyu Chen, Guofeng Zhang, Yu-Cheng Chou, Jieneng Chen, Celso de Melo, and Alan Yuille.
\newblock 3dsrbench: A comprehensive 3d spatial reasoning benchmark.
\newblock In \emph{Proceedings of the IEEE/CVF International Conference on Computer Vision}, pages 6924--6934, 2025{\natexlab{a}}.

\bibitem[Ma et~al.(2025{\natexlab{b}})Ma, Chou, Liu, Wang, de~Melo, Xie, and Yuille]{ma2025spatialreasoner}
Wufei Ma, Yu-Cheng Chou, Qihao Liu, Xingrui Wang, Celso~M de Melo, Jianwen Xie, and Alan Yuille.
\newblock Spatialreasoner: Towards explicit and generalizable 3d spatial reasoning.
\newblock In \emph{The Annual Conference on Neural Information Processing Systems}, 2025{\natexlab{b}}.

\bibitem[Meng et~al.(2025)Meng, Wang, Li, Lu, Tian, Liao, Zhu, Dai, Qiao, Luo, et~al.]{meng2024mmiu}
Fanqing Meng, Jin Wang, Chuanhao Li, Quanfeng Lu, Hao Tian, Jiaqi Liao, Xizhou Zhu, Jifeng Dai, Yu Qiao, Ping Luo, et~al.
\newblock Mmiu: Multimodal multi-image understanding for evaluating large vision-language models.
\newblock In \emph{The International Conference on Learning Representations}, 2025.

\bibitem[Ogezi and Shi(2025)]{ogezi2025spare}
Michael Ogezi and Freda Shi.
\newblock Spare: Enhancing spatial reasoning in vision-language models with synthetic data.
\newblock In \emph{Proceedings of the Annual Meeting of the Association for Computational Linguistics (Volume 1: Long Papers)}, pages 7855--7875, 2025.

\bibitem[Qi et~al.(2025)]{qi2025beyond}
Yuan Qi et~al.
\newblock Beyond semantics: Rediscovering spatial awareness in vision-language models.
\newblock \emph{arXiv preprint arXiv:2503.17349}, 2025.

\bibitem[Rajabi et~al.(2024)]{rajabi2024towards}
Reza Rajabi et~al.
\newblock Towards grounded visual spatial reasoning in multi-modal vision language models.
\newblock In \emph{ICLR Workshop}, 2024.

\bibitem[Ray et~al.(2024)Ray, Duan, Tan, Bashkirova, Hendrix, Ehsani, Kembhavi, Plummer, Krishna, Zeng, et~al.]{SAT}
Arijit Ray, Jiafei Duan, Reuben Tan, Dina Bashkirova, Rose Hendrix, Kiana Ehsani, Aniruddha Kembhavi, Bryan~A Plummer, Ranjay Krishna, Kuo-Hao Zeng, et~al.
\newblock Sat: Spatial aptitude training for multimodal language models.
\newblock \emph{arXiv preprint arXiv:2412.07755}, 3, 2024.

\bibitem[Ren et~al.(2025)Ren, Shen, Huang, Ling, Lu, Nimier-David, Müller, Keller, Fidler, and Gao]{ren2025gen3c}
Xuanchi Ren, Tianchang Shen, Jiahui Huang, Huan Ling, Yifan Lu, Merlin Nimier-David, Thomas Müller, Alexander Keller, Sanja Fidler, and Jun Gao.
\newblock Gen3c: 3d-informed world-consistent video generation with precise camera control.
\newblock In \emph{Proceedings of the IEEE/CVF Conference on Computer Vision and Pattern Recognition}, 2025.

\bibitem[Sharma et~al.(2026)Sharma, Hogervorst, Mackey, Heeger, and Martiniani]{sharma2026crossviewworldmodels}
Rishabh Sharma, Gijs Hogervorst, Wayne~E. Mackey, David~J. Heeger, and Stefano Martiniani.
\newblock Cross-view world models, 2026.

\bibitem[Su et~al.(2025)Su, Liu, Wang, Wei, Kang, Ruan, and Zhu]{su2025reactivecognitivebraininspired}
Hang Su, Songming Liu, Liyuan Wang, Xingxing Wei, Caixin Kang, Shouwei Ruan, and Qihui Zhu.
\newblock From reactive to cognitive: brain-inspired spatial intelligence for embodied agents, 2025.

\bibitem[Tian et~al.(2025)Tian, Liu, Wang, Tao, Li, Yang, Peng, Xue, Li, Wang, Wang, Xu, Li, Chen, Long, Tan, Yuan, Lin, Wang, Wu, Jiang, Wang, Zhang, Dai, Li, Zhang, Liu, Yang, Wang, Cui, Bai, Xiong, Deng, Lu, Yu, Zhou, Kong, Zhou, Song, He, Huang, Wu, Xu, Liu, Min, Wu, Wang, Wang, Li, Yu, Zhong, and Zhou]{tian2025hunyuanvideosystematicframework}
Qi Tian, Songtao Liu, Di Wang, Yangyu Tao, Yang Li, Fang Yang, Yuanbo Peng, Jinbao Xue, Xin Li, Hongfa Wang, Andong Wang, Zunnan Xu, Shuai Li, Yi Chen, Yanxin Long, Hao Tan, Junkun Yuan, Qin Lin, Kai Wang, Bo Wu, Jie Jiang, Weiyan Wang, Zijian Zhang, Zuozhuo Dai, Changlin Li, Jianwei Zhang, Yuhong Liu, Yong Yang, Hongmei Wang, Yutao Cui, Jiawang Bai, Jiangfeng Xiong, Xinchi Deng, Qinglin Lu, Wenqing Yu, Jin Zhou, Weijie Kong, Zixiang Zhou, Jacob Song, Zhiyu He, Duojun Huang, Jianbing Wu, Zhiyong Xu, Mengyang Liu, Rox Min, Kathrina Wu, Aladdin Wang, Joey Wang, Pengyu Li, Zhentao Yu, Caesar Zhong, and Dax Zhou.
\newblock Hunyuanvideo: A systematic framework for large video generative models, 2025.

\bibitem[Wang et~al.(2025)Wang, Tan, Zhu, Yang, Yang, Wang, Kolobov, Gao, and Gong]{wang2025site}
Wenqi Wang, Reuben Tan, Pengyue Zhu, Jianwei Yang, Zhengyuan Yang, Lijuan Wang, Andrey Kolobov, Jianfeng Gao, and Boqing Gong.
\newblock Site: towards spatial intelligence thorough evaluation.
\newblock In \emph{Proceedings of the IEEE/CVF International Conference on Computer Vision}, pages 9058--9069, 2025.

\bibitem[Wang et~al.(2026)Wang, Lin, Yoon, Cho, Zhang, and Bansal]{wang2026anchorweaveworldconsistentvideogeneration}
Zun Wang, Han Lin, Jaehong Yoon, Jaemin Cho, Yue Zhang, and Mohit Bansal.
\newblock Anchorweave: World-consistent video generation with retrieved local spatial memories, 2026.

\bibitem[Wu et~al.(2025{\natexlab{a}})Wu, Liu, Hung, and Duan]{wu2025spatialmllmboostingmllmcapabilities}
Diankun Wu, Fangfu Liu, Yi-Hsin Hung, and Yueqi Duan.
\newblock Spatial-mllm: Boosting mllm capabilities in visual-based spatial intelligence, 2025{\natexlab{a}}.

\bibitem[Wu et~al.(2025{\natexlab{b}})Wu, Huang, Chen, Zhang, Wang, and Xie]{wu2025spatialscore}
Haoning Wu, Xiao Huang, Yaohui Chen, Ya Zhang, Yanfeng Wang, and Weidi Xie.
\newblock Spatialscore: Towards unified evaluation for multimodal spatial understanding.
\newblock \emph{arXiv preprint arXiv:2505.17012}, 2025{\natexlab{b}}.

\bibitem[Wu et~al.(2026)Wu, Cai, Li, Ge, Sun, Yuan, and Wang]{wu2026camreasonerreinforcingcameramovement}
Hang Wu, Yujun Cai, Zehao Li, Haonan Ge, Bowen Sun, Junsong Yuan, and Yiwei Wang.
\newblock Camreasoner: Reinforcing camera movement understanding via structured spatial reasoning, 2026.

\bibitem[Wu et~al.(2025{\natexlab{c}})Wu, Guan, Feng, Liu, Wu, Wang, Wu, and Tan]{wu2025reinforcingspatialreasoningvisionlanguage}
Junfei Wu, Jian Guan, Kaituo Feng, Qiang Liu, Shu Wu, Liang Wang, Wei Wu, and Tieniu Tan.
\newblock Reinforcing spatial reasoning in vision-language models with interwoven thinking and visual drawing, 2025{\natexlab{c}}.

\bibitem[Wu et~al.(2025{\natexlab{d}})Wu, Yang, Zhou, Fang, Song, Sun, and Ji]{wu2025groundedchainofthoughtmultimodallarge}
Qiong Wu, Xiangcong Yang, Yiyi Zhou, Chenxin Fang, Baiyang Song, Xiaoshuai Sun, and Rongrong Ji.
\newblock Grounded chain-of-thought for multimodal large language models, 2025{\natexlab{d}}.

\bibitem[Xie et~al.(2025)Xie, Liu, Krishna, Fei-Fei, Wu, Wang, Li, Zhang, Zhang, Wang, Wang, Chandrasegaran, Yin, and Zhang]{yin2025mindcube}
Saining Xie, Han Liu, Ranjay Krishna, Li Fei-Fei, Jiajun Wu, Zihan Wang, Manling Li, Jieyu Zhang, Jianshu Zhang, Qineng Wang, Kangrui Wang, Keshigeyan Chandrasegaran, Baiqiao Yin, and Pingyue Zhang.
\newblock Spatial mental modeling from limited views, 2025.

\bibitem[Xu et~al.(2023)Xu, Xie, Qin, Tao, and Wang]{xu2023parameter}
Lingling Xu, Haoran Xie, Si-Zhao~Joe Qin, Xiaohui Tao, and Fu~Lee Wang.
\newblock Parameter-efficient fine-tuning methods for pretrained language models: A critical review and assessment.
\newblock \emph{arXiv preprint arXiv:2312.12148}, 2023.

\bibitem[Yang et~al.(2025{\natexlab{a}})Yang, Du, Gan, Zhou, Tan, Yang, Zhang, and Liu]{yang2025mindjourney}
Jianwei Yang, Yilun Du, Chuang Gan, Siyuan Zhou, Reuben Tan, Yuncong Yang, Zheyuan Zhang, and Jiageng Liu.
\newblock Mindjourney: Test-time scaling with world models for spatial reasoning, 2025{\natexlab{a}}.

\bibitem[Yang et~al.(2025{\natexlab{b}})Yang, Yang, Gupta, Han, Fei-Fei, and Xie]{yang2025thinking}
Jihan Yang, Shusheng Yang, Anjali~W Gupta, Rilyn Han, Li Fei-Fei, and Saining Xie.
\newblock Thinking in space: How multimodal large language models see, remember, and recall spaces.
\newblock In \emph{Proceedings of the Computer Vision and Pattern Recognition Conference}, pages 10632--10643, 2025{\natexlab{b}}.

\bibitem[Yang et~al.(2025{\natexlab{c}})Yang, Xu, Xie, Yang, Li, Lin, Zhu, Chen, Duan, Yue, et~al.]{yang2025mmsi}
Sihan Yang, Runsen Xu, Yiman Xie, Sizhe Yang, Mo Li, Jingli Lin, Chenming Zhu, Xiaochen Chen, Haodong Duan, Xiangyu Yue, et~al.
\newblock Mmsi-bench: A benchmark for multi-image spatial intelligence.
\newblock \emph{arXiv preprint arXiv:2505.23764}, 2025{\natexlab{c}}.

\bibitem[Yang et~al.(2025{\natexlab{d}})Yang, Yang, Huang, Brown, Yang, Yu, Tong, Zheng, Xu, Wang, et~al.]{yang2025cambrian}
Shusheng Yang, Jihan Yang, Pinzhi Huang, Ellis Brown, Zihao Yang, Yue Yu, Shengbang Tong, Zihan Zheng, Yifan Xu, Muhan Wang, et~al.
\newblock Cambrian-s: Towards spatial supersensing in video.
\newblock \emph{arXiv preprint arXiv:2511.04670}, 2025{\natexlab{d}}.

\bibitem[Yang et~al.(2026)Yang, Fan, Shi, Peng, Wang, and Zhang]{yang2026neoverse}
Yuxue Yang, Lue Fan, Ziqi Shi, Junran Peng, Feng Wang, and Zhaoxiang Zhang.
\newblock Neoverse: Enhancing 4d world model with in-the-wild monocular videos.
\newblock \emph{arXiv preprint arXiv:2601.00393}, 2026.

\bibitem[Yeh et~al.(2025)Yeh, Wang, Tong, Cheng, Wang, Chu, Zhai, Chen, Gao, and Ma]{yeh2025seeingperspectiveevaluatingmultiview}
Chun-Hsiao Yeh, Chenyu Wang, Shengbang Tong, Ta-Ying Cheng, Ruoyu Wang, Tianzhe Chu, Yuexiang Zhai, Yubei Chen, Shenghua Gao, and Yi Ma.
\newblock Seeing from another perspective: Evaluating multi-view understanding in mllms.
\newblock In \emph{Neural Information Processing Systems}, 2025.

\bibitem[Yu et~al.(2025)Yu, Chen, Ju, Jia, Zhang, Huang, Wu, Cui, Ran, Zhang, et~al.]{yu2025far}
Songsong Yu, Yuxin Chen, Hao Ju, Lianjie Jia, Fuxi Zhang, Shaofei Huang, Yuhan Wu, Rundi Cui, Binghao Ran, Zaibin Zhang, et~al.
\newblock How far are vlms from visual spatial intelligence? a benchmark-driven perspective.
\newblock \emph{arXiv preprint arXiv:2509.18905}, 2025.

\bibitem[Zhan et~al.(2025)Zhan, Huang, Sun, Fu, Ma, Cao, Jia, Lin, Yin, BAI, Ouyang, Li, Guo, and Guo]{zhan2025actial}
Xiaoyu Zhan, Wenxuan Huang, Hao Sun, Xinyu Fu, Changfeng Ma, Shaosheng Cao, Bohan Jia, Shaohui Lin, Zhenfei Yin, LEI BAI, Wanli Ouyang, Yuanqi Li, Jie Guo, and Yanwen Guo.
\newblock Actial: Activate spatial reasoning ability of multimodal large language models.
\newblock In \emph{The Annual Conference on Neural Information Processing Systems}, 2025.

\bibitem[Zhang et~al.(2025{\natexlab{a}})Zhang, Chen, Zhou, Xu, Huang, Mei, Chen, Yuan, Cai, Huang, et~al.]{zhang2025flatland}
Jiahui Zhang, Yurui Chen, Yanpeng Zhou, Yueming Xu, Ze Huang, Jilin Mei, Junhui Chen, Yu-Jie Yuan, Xinyue Cai, Guowei Huang, et~al.
\newblock From flatland to space: Teaching vision-language models to perceive and reason in 3d.
\newblock \emph{arXiv preprint arXiv:2503.22976}, 2025{\natexlab{a}}.

\bibitem[Zhang et~al.(2025{\natexlab{b}})Zhang, Ren, Xu, Huang, Zhang, Xu, Huang, and Zhi]{zhang2025mulset}
Jiajun Zhang, Shuo Ren, Wang Xu, Yibin Huang, Wanyue Zhang, Yangbin Xu, JingJing Huang, and Helu Zhi.
\newblock Why do mllms struggle with spatial understanding? a systematic analysis from data to architecture, 2025{\natexlab{b}}.

\bibitem[Zhang et~al.(2025{\natexlab{c}})Zhang, Ge, Chi, Zhang, Shi, Dong, Han, and Zhang]{zhang2025worldmodelsbenefitvlms}
Kevin Zhang, Kuangzhi Ge, Xiaowei Chi, Renrui Zhang, Shaojun Shi, Zhen Dong, Sirui Han, and Shanghang Zhang.
\newblock Can world models benefit vlms for world dynamics?, 2025{\natexlab{c}}.

\bibitem[Zhang et~al.(2026)Zhang, Wu, Jia, Wang, Zhang, Li, Ran, Zhang, Sun, Yin, Wang, and Lu]{zhang2026think3dthinkingspacespatial}
Zaibin Zhang, Yuhan Wu, Lianjie Jia, Yifan Wang, Zhongbo Zhang, Yijiang Li, Binghao Ran, Fuxi Zhang, Zhuohan Sun, Zhenfei Yin, Lijun Wang, and Huchuan Lu.
\newblock Think3d: Thinking with space for spatial reasoning, 2026.

\bibitem[Zheng et~al.(2025)Zheng, Dongfang, Jiang, Zheng, Guo, Zhang, Albanese, Yang, Ma, Zhang, et~al.]{zheng2025multimodal}
Xu Zheng, Zihao Dongfang, Lutao Jiang, Boyuan Zheng, Yulong Guo, Zhenquan Zhang, Giuliano Albanese, Runyi Yang, Mengjiao Ma, Zixin Zhang, et~al.
\newblock Multimodal spatial reasoning in the large model era: A survey and benchmarks.
\newblock \emph{arXiv preprint arXiv:2510.25760}, 2025.

\bibitem[Zhou et~al.(2025)Zhou, Gao, Voleti, Vasishta, Yao, Boss, Torr, Rupprecht, and Jampani]{zhou2025stable}
Jensen Zhou, Hang Gao, Vikram Voleti, Aaryaman Vasishta, Chun-Han Yao, Mark Boss, Philip Torr, Christian Rupprecht, and Varun Jampani.
\newblock Stable virtual camera: Generative view synthesis with diffusion models.
\newblock In \emph{Proceedings of the IEEE/CVF International Conference on Computer Vision}, pages 12405--12414, 2025.

\bibitem[Zhu et~al.(2025)Zhu, Wang, Chen, Liu, Ye, Gu, Tian, Duan, Su, Shao, Gao, Cui, Wang, Cao, Liu, Wei, Zhang, Wang, Xu, Li, Wang, Deng, Li, He, Jiang, Luo, Wang, He, Shi, Zhang, Shao, He, Xiong, Qu, Sun, Jiao, Lv, Wu, Zhang, Deng, Ge, Chen, Wang, Dou, Lu, Zhu, Lu, Lin, Qiao, Dai, and Wang]{zhu2025internvl3exploringadvancedtraining}
Jinguo Zhu, Weiyun Wang, Zhe Chen, Zhaoyang Liu, Shenglong Ye, Lixin Gu, Hao Tian, Yuchen Duan, Weijie Su, Jie Shao, Zhangwei Gao, Erfei Cui, Xuehui Wang, Yue Cao, Yangzhou Liu, Xingguang Wei, Hongjie Zhang, Haomin Wang, Weiye Xu, Hao Li, Jiahao Wang, Nianchen Deng, Songze Li, Yinan He, Tan Jiang, Jiapeng Luo, Yi Wang, Conghui He, Botian Shi, Xingcheng Zhang, Wenqi Shao, Junjun He, Yingtong Xiong, Wenwen Qu, Peng Sun, Penglong Jiao, Han Lv, Lijun Wu, Kaipeng Zhang, Huipeng Deng, Jiaye Ge, Kai Chen, Limin Wang, Min Dou, Lewei Lu, Xizhou Zhu, Tong Lu, Dahua Lin, Yu Qiao, Jifeng Dai, and Wenhai Wang.
\newblock Internvl3: Exploring advanced training and test-time recipes for open-source multimodal models, 2025.

\end{thebibliography}
}

\newpage
\appendix
\section{Detailed Data Construction}
\label{sec:appendix-data-construction}

\paragraph{Anchor RGB sources.}
We construct World2VLM supervision from two complementary anchor-image pools. For real-scene supervision, we use ScanNet RGB frames. In the current release manifest used by our SVC pipeline, this source contributes 6,803 anchor images. For simulated supervision, we use a selected MulSeT RGB pool, which contributes 2,000 anchor images. Each manifest entry is treated as one anchor scene instance: the generator takes a single source RGB image, samples an egocentric motion program, synthesizes a trajectory of future views, and then converts the resulting trajectory bundle into one or more QA-style supervision records.

\paragraph{Trajectory generation with SVC.}
For the SVC teacher, trajectory generation uses a camera-conditioned novel-view pipeline with resolution $576\times576$, guidance scale 4.0, camera scale 2.0, and 50 denoising steps. The action space contains six single-step egocentric actions: \textit{forward}, \textit{backward}, \textit{shift-left}, \textit{shift-right}, \textit{turn-left}, and \textit{turn-right}. In addition, we use three short multi-step presets: \textit{forward-then-turn-left}, \textit{turn-right-then-forward}, and \textit{shift-left-then-forward-then-turn-right}. Translation magnitudes are sampled on a discrete 0.1\,m grid and rotation magnitudes on a discrete 10$^\circ$ grid. Single-step translations are sampled up to 6.0\,m, single-step rotations up to 100$^\circ$, and long motions are decomposed into small per-frame increments so that trajectories remain temporally smooth. Each generated trajectory contains at least 8 frames, and the trajectory metadata stores both step-wise actions and cumulative action prefixes.

For prompt construction, we use a max-displacement policy: the anchor frame is paired with the farthest valid synthesized endpoint rather than a near-neighbor frame. This design is used for both motion-only and object-grounded prompts so that supervision emphasizes non-trivial viewpoint change instead of local appearance continuity.

\paragraph{Trajectory generation with HY-WorldPlay.}
When we replace the teacher with HY-WorldPlay, we keep the same anchor-image sources, downstream templates, and train/validation split protocol, and only swap the trajectory-generation backend. HY-WorldPlay is instantiated as an action-conditioned video world model using few-step bf16 inference at 480p ($832\times480$, aspect ratio 16:9). Its motion space again covers six single-step actions plus short multi-step presets, with translation magnitudes quantized in 0.1\,m units and rotations in 10$^\circ$ units. Relative to SVC, HY-WorldPlay therefore changes the world-model prior and rendering mechanism, but not the downstream QA construction procedure.

\paragraph{Prompt construction.}
After trajectory synthesis, we convert each trajectory bundle into two supervision families. Motion-only prompts (A1--A4) are produced directly from the image pair and the stored action prefix. Specifically, A1 predicts translation distance, A2 predicts rotation angle, A3 predicts a 2--3 step ordered action sequence, and A4 verifies whether a proposed action sequence matches the source-target transition. Detector-grounded prompts (D1--D4) are built by running object detection and tracking on the generated trajectory frames, caching the detections, normalizing boxes to integer coordinates in $[0,1000]$, and instantiating task-specific templates. D1 predicts the post-action target box, D2 predicts whether the object remains visible, D3 infers a 1--2 step action sequence from matched boxes across views, and D4 judges whether two boxes refer to the same physical object instance.

\paragraph{Filtering and quality control.}
We apply filtering at both the object level and the record level. For object-grounded prompts, we keep only detections that satisfy confidence $\ge 0.3$, NMS IoU threshold 0.5, area ratio in $[0.01, 0.6]$, and a minimum 1\% border margin from all image boundaries. When multiple candidates remain, we keep the highest-confidence valid track-consistent match. At the record level, prompt builders reject malformed outputs, invalid boxes, and action/trajectory mismatches. In the balanced subset used for GRPO data construction, 112 malformed records were removed by this validator, all due to invalid D3 action formatting. In practice, most pruning happens after generation during prompt validation and balancing rather than by discarding large numbers of trajectories through aggressive temporal rejection.

\paragraph{Raw prompt pools.}
The four SVC prompt pools used to build the final supervised and reinforcement-learning data are summarized in Table~\ref{tab:appendix-data-pools}. ScanNet contributes both motion-only and detector-grounded records, as does MulSeT. Before final packaging, these four raw pools contain 103,339 candidate QA pairs in total: 52,798 ScanNet motion-only records, 13,482 ScanNet detector-grounded records, 23,675 MulSeT motion-only records, and 13,384 MulSeT detector-grounded records.

\begin{table}[t]
\centering
\small
\setlength{\tabcolsep}{5pt}
\begin{tabular}{lcccc}
\toprule
\textbf{Source} & \textbf{Anchor images} & \textbf{A1--A4} & \textbf{D1--D4} & \textbf{Final SFT} \\
\midrule
ScanNet & 6,803 & 52,798 & 13,482 & 66,174 \\
MulSeT & 2,000 & 23,675 & 13,384 & 36,923 \\
\midrule
Total & 8,803 & 76,473 & 26,866 & 103,097 \\
\bottomrule
\end{tabular}
\caption{Source-level statistics for the World2VLM training data. ``A1--A4'' denotes motion-only prompts and ``D1--D4'' denotes detector-grounded prompts before final SFT packaging. The final SFT file contains 103,097 records, which we refer to as ``100K'' in the main text for readability.}
\label{tab:appendix-data-pools}
\end{table}

\paragraph{Final SFT composition.}
The final supervised training file contains 103,097 records in total. Aggregated by supervision direction, it contains 61,427 inverse examples and 41,670 forward examples. Aggregated by task family, it contains 76,473 motion-only examples (A1--A4) and 26,624 detector-grounded examples (D1--D4). Table~\ref{tab:appendix-task-counts} gives the per-task breakdown. Because object-grounded tasks require valid tracked detections and stable cross-view matches, the final task distribution is not perfectly uniform, even though the overall construction procedure attempts to balance task family, trajectory group, and source domain.

\begin{table}[t]
\centering
\small
\setlength{\tabcolsep}{8pt}
\begin{tabular}{lcc}
\toprule
\textbf{Task} & \textbf{Count} & \textbf{Family} \\
\midrule
A1 & 24,788 & Inverse / motion-only \\
A2 & 13,903 & Inverse / motion-only \\
A3 & 18,891 & Inverse / motion-only \\
A4 & 18,891 & Forward / motion-only \\
D1 & 4,158 & Forward / detector-grounded \\
D2 & 13,039 & Forward / detector-grounded \\
D3 & 3,845 & Inverse / detector-grounded \\
D4 & 5,582 & Forward / detector-grounded \\
\bottomrule
\end{tabular}
\caption{Per-task counts in the final World2VLM supervised training file.}
\label{tab:appendix-task-counts}
\end{table}

\paragraph{GRPO refinement set.}
For reinforcement learning, we construct a separate balanced subset from the same four raw SVC pools. The resulting GRPO set contains 1,000 examples. This set is balanced by task type (125 examples for each of A1--A4 and D1--D4) and by source bucket (250 examples each from \textit{scannet\_detect}, \textit{scannet\_undetect}, \textit{mulset\_detect}, and \textit{mulset\_undetect}), while also maintaining near-uniform coverage over the major trajectory groups. As described in the main paper, we then reserve a small validation split from this balanced refinement pool for checkpoint selection and reward debugging.

\section{Task Templates and Output Formats}
\label{sec:appendix-task-templates}

\paragraph{Canonical template set.}
Table-free summaries in the main paper describe the semantics of A1--A4 and D1--D4, but reviewers may still want to see the exact prompt surface form and target answer style. We therefore present the canonical template family used for dataset construction. For readability in the two-column format, we show one canonical prompt form and one concrete input-output example per task. Some object-grounded tasks also admit light wording variants such as multiple-choice or short-sentence versions, but the canonical forms below capture the supervision pattern used throughout the dataset.

\paragraph{A1: Translation distance estimation.}\mbox{}\\
This is an inverse motion-prediction task over an image pair. The model must infer the signed translation needed to transform the first view into the second view.

\textbf{Prompt template.}
\begin{quote}
\small
\texttt{<image><image>How many meters did the camera move to get the second image?}\\
\texttt{Answer as: move \{DIR\} X meters.}
\end{quote}

\textbf{Target output format.} \texttt{move \{forward|backward|left|right\} \{DIST\} meters}

\textbf{Example.}
\begin{quote}
\small
\texttt{User: <image><image>How many meters did the camera move to get the second image? Answer as: move forward X meters.}\\
\texttt{Assistant: move forward 4.3 meters}
\end{quote}

\paragraph{A2: Rotation angle estimation.}\mbox{}\\
This is an inverse motion-prediction task over an image pair. The model must infer the camera's turn direction and rotation magnitude from the source-target transition.

\textbf{Prompt template.}
\begin{quote}
\small
\texttt{<image><image>How many degrees did the camera turn to get the second image?}\\
\texttt{Answer as: turn \{TURN\_DIR\} X degrees.}
\end{quote}

\textbf{Target output format.} \texttt{turn \{left|right\} \{ANGLE\} degrees}

\textbf{Example.}
\begin{quote}
\small
\texttt{User: <image><image>How many degrees did the camera turn to get the second image? Answer as: turn left X degrees.}\\
\texttt{Assistant: turn left 100 degrees}
\end{quote}

\paragraph{A3: Multi-step action-sequence prediction.}\mbox{}\\
This is an inverse structured-prediction task. Given a source-target image pair linked by a short composed trajectory, the model must reconstruct the ordered action sequence rather than a single scalar motion value.

\textbf{Prompt template.}
\begin{quote}
\small
\texttt{<image><image>To move from the first image to the second image, the camera used 2 or 3 actions in order.}\\
\texttt{Write the full action sequence using ';' as a separator. Allowed actions:}\\
\texttt{- move forward/backward/left/right X meters}\\
\texttt{- turn left/right X degrees}
\end{quote}

\textbf{Target output format.} \texttt{action\_1; action\_2} or \texttt{action\_1; action\_2; action\_3}, where each action is one of \texttt{move ... X meters} or \texttt{turn ... X degrees}

\textbf{Example.}
\begin{quote}
\small
\texttt{User: <image><image>To move from the first image to the second image, the camera used 2 or 3 actions in order. Write the full action sequence using ';' as a separator.}\\
\texttt{Assistant: move forward 1.8 meters; turn left 50 degrees}
\end{quote}

\paragraph{A4: Action verification.}\mbox{}\\
This is a forward verification task. Instead of predicting the action from scratch, the model must judge whether a candidate motion description is consistent with the image pair.

\textbf{Prompt template.}
\begin{quote}
\small
\texttt{<image><image>True or false: the camera did "\{CLAIM\_ACTION\}" to get the second image.}
\end{quote}

\textbf{Target output format.} \texttt{true} or \texttt{false}

\textbf{Example.}
\begin{quote}
\small
\texttt{User: <image><image>True or false: the camera did "move left 0.8 meters, move forward 2.8 meters and turn right 50 degrees" to get the second image.}\\
\texttt{Assistant: true}
\end{quote}

\paragraph{D1: Post-action object localization.}\mbox{}\\
This is a forward object-grounding task over a paired-view transition. The model is given an object box in the source image and must predict the corresponding box after the specified action.

\textbf{Prompt template.}
\begin{quote}
\small
\texttt{<image><image>In the first image, the \{OBJECT\} is at bbox \{BBOX\_1\}.}\\
\texttt{Bboxes use normalized integer coordinates in [0,1000]. After the camera does "\{ACTION\_SEQ\}",}\\
\texttt{give the bbox of the same \{OBJECT\} in the second image. Answer with bbox [x1, y1, x2, y2] only.}
\end{quote}

\textbf{Target output format.} \texttt{[x1, y1, x2, y2]} with integer coordinates in \texttt{[0,1000]}

\textbf{Example.}
\begin{quote}
\small
\texttt{User: <image><image>In the first image, the microwave is at bbox [94, 423, 322, 552]. Bboxes use normalized integer coordinates in [0,1000]. After the camera does "move backward 4.5 meters", give the bbox of the same microwave in the second image. Answer with bbox [x1, y1, x2, y2] only.}\\
\texttt{Assistant: [48, 558, 226, 681]}
\end{quote}

\paragraph{D2: Post-action visibility judgment.}\mbox{}\\
This is a forward object-centric reasoning task. The model receives an object box in the source image and must determine whether the same object remains visible after the action.

\textbf{Prompt template.}
\begin{quote}
\small
\texttt{<image>In the image, the \{OBJECT\} is at bbox \{BBOX\_1\}. After the camera does "\{ACTION\_SEQ\}",}\\
\texttt{does this \{OBJECT\} disappear from view? Answer: yes or no.}
\end{quote}

\textbf{Target output format.} \texttt{yes} or \texttt{no}

\textbf{Example.}
\begin{quote}
\small
\texttt{User: <image>In the image, the chair is at bbox [626, 209, 915, 675]. After the camera does "move forward 5.5 meters", does this chair disappear from view? Answer: yes or no.}\\
\texttt{Assistant: yes}
\end{quote}

\paragraph{D3: Box-guided action-sequence inference.}\mbox{}\\
This is an inverse paired-view matching task. The model is told that the two boxes correspond to the same physical object instance and must infer the short camera action sequence that explains their relative displacement.

\textbf{Prompt template.}
\begin{quote}
\small
\texttt{<image><image>The \{OBJECT\} in the first image (bbox \{BBOX\_1\}) and the \{OBJECT\} in the second image (bbox \{BBOX\_2\}) are the same physical object.}\\
\texttt{The camera used 1 or 2 actions in order. Write the full action sequence using ';' as a separator. Allowed actions:}\\
\texttt{- move forward/backward/left/right X meters}\\
\texttt{- turn left/right X degrees}
\end{quote}

\textbf{Target output format.} \texttt{action\_1} or \texttt{action\_1; action\_2}, using the same action vocabulary as A3

\textbf{Example.}
\begin{quote}
\small
\texttt{User: <image><image>The cup in the first image (bbox [412, 557, 523, 700]) and the cup in the second image (bbox [603, 368, 814, 877]) are the same physical object. The camera used 1 or 2 actions in order. Write the full action sequence using ';' as a separator.}\\
\texttt{Assistant: move forward 3 meters; turn left 50 degrees}
\end{quote}

\paragraph{D4: Cross-view object-identity verification.}\mbox{}\\
This is a forward paired-view verification task. Given a source box, a target box, and the applied action, the model must judge whether the two boxes correspond to the same object instance.

\textbf{Prompt template.}
\begin{quote}
\small
\texttt{<image><image>In the first image, the \{OBJECT\} is at bbox \{BBOX\_1\}. After the camera does "\{ACTION\_SEQ\}",}\\
\texttt{the second image shows a \{OBJECT\} at bbox \{BBOX\_2\}. Are these the same physical object instance? Answer: yes or no.}
\end{quote}

\textbf{Target output format.} \texttt{yes} or \texttt{no}

\textbf{Example.}
\begin{quote}
\small
\texttt{User: <image><image>In the first image, the chair is at bbox [70, 519, 404, 986]. After the camera does "move left 5 meters", the second image shows a chair at bbox [384, 525, 684, 978]. Are these the same physical object instance? Answer: yes or no.}\\
\texttt{Assistant: no}
\end{quote}

\paragraph{Interpretation.}
Taken together, these templates cover three supervision modes. A1, A2, A3, and D3 are inverse tasks that infer motion from visual change. A4 and D4 are verification tasks that judge whether a proposed action or identity relation is compatible with the transition. D1 and D2 are forward consequence-prediction tasks that forecast object state after an action. This explicit split is important for interpreting World2VLM: the model is not trained on a single spatial QA format, but on a mixed supervision set that spans scalar regression, sequence prediction, grounding, visibility prediction, and paired-view verification.

\section{Reward Design for GRPO}
\label{sec:appendix-grpo-reward}

\paragraph{Implementation scope.}
This section documents the task-aware reward used in the multitype GRPO stage. The reported run uses a balanced 1K refinement pool, 4 rollout samples per prompt, actor/critic learning rates of $2\times 10^{-6}$ and $1\times 10^{-6}$, actor/critic global batch sizes of 2, fixed KL coefficient 0.01, and one training epoch. The reward is evaluated from the model response together with the task identifier and reference answer, and it returns a scalar task reward together with interpretable sub-scores for analysis.

\paragraph{Common preprocessing and score range.}
For action and binary tasks, the implementation lowercases the response, strips trailing punctuation, and collapses repeated whitespace before parsing. For D1, the raw response is kept so that partial bounding-box formatting can still receive partial credit. All component rewards lie in $[0,1]$. The only negative case is an explicit overlength penalty: if the raw response exceeds 200 characters, the sample immediately receives the minimum task reward and zero format credit, without evaluating any task-specific sub-scores. This rule is important in our setting because the rollout prompt asks the model to produce intermediate reasoning, and the penalty discourages long malformed chains of thought that never collapse to a clean final answer.

\paragraph{Shared numeric precision term.}
For action-valued tasks, magnitude accuracy is measured by a piecewise-linear function:
\begin{equation}
s_{\mathrm{num}}(\hat{v}, v^\star; \tau)=
\begin{cases}
1, & |\hat{v}-v^\star| \le \tau_{\mathrm{low}},\\
0, & |\hat{v}-v^\star| \ge \tau_{\mathrm{high}},\\
1-\dfrac{|\hat{v}-v^\star|-\tau_{\mathrm{low}}}{\tau_{\mathrm{high}}-\tau_{\mathrm{low}}}, & \text{otherwise}.
\end{cases}
\end{equation}
For translation magnitudes, $(\tau_{\mathrm{low}}, \tau_{\mathrm{high}})=(0.5\text{ m}, 5.0\text{ m})$; for turn angles, $(\tau_{\mathrm{low}}, \tau_{\mathrm{high}})=(5^\circ, 90^\circ)$. Thus $r_{\text{num}}$ equals 1 for small numeric error, decays linearly in the mid-error regime, and saturates at 0 for large mismatch.

\paragraph{Task-specific instantiations.}
Table~\ref{tab:appendix-grpo-weights} summarizes which reward terms are active for each task family and how they are weighted in the implementation.

\begin{table*}[t]
\centering
\small
\setlength{\tabcolsep}{6pt}
\begin{tabular}{lccccccl}
\toprule
\textbf{Task(s)} & $\alpha_{\text{fmt}}$ & $\alpha_{\text{sem}}$ & $\alpha_{\text{num}}$ & $\alpha_{\text{geo}}$ & $\alpha_{\text{ord}}$ & \textbf{Extra term} & \textbf{Implementation form} \\
\midrule
A1, A2 & 0.10 & 0.35 & 0.55 & 0 & 0 & none & single-step motion reward \\
A3, D3 & 0.10 & 0.25 & 0.30 & 0 & 0.35 & $-0.03$ per extra action & sequence reward \\
A4, D2, D4 & 0.20 & 0.80 & 0 & 0 & 0 & none & binary reward \\
D1 & 0.20 & 0 & 0 & 0.65 & 0 & $0.15\,r_{\text{valid}}$ & geometric box reward \\
\bottomrule
\end{tabular}
\caption{Task-specific reward coefficients used by the GRPO implementation. Eq.~(4) in the main text is therefore a shorthand summary: the active terms and their coefficients are instantiated differently for different task types. D1 additionally includes an auxiliary box-validity term $r_{\text{valid}}$.}
\label{tab:appendix-grpo-weights}
\end{table*}

\paragraph{A1/A2: single-step motion rewards.}
For translation-distance (A1) and turn-angle (A2) prediction, the reward parser extracts all action strings matching
\texttt{move forward/backward/left/right X meters} or
\texttt{turn left/right X degrees} from the response. Let $(a^\star, v^\star)$ denote the ground-truth action type and magnitude, and let $\{(\hat{a}_j, \hat{v}_j)\}_j$ be the parsed candidates. The component rewards are
\begin{equation}
\begin{aligned}
r_{\text{fmt}} &= \mathbb{1}[|\hat{\mathcal{A}}|>0], \qquad
r_{\text{sem}} = \max_j \mathbb{1}[\hat{a}_j=a^\star], \\
r_{\text{num}} &= \max_j \mathbb{1}[\hat{a}_j=a^\star]\,
s_{\mathrm{num}}(\hat{v}_j, v^\star; a^\star).
\end{aligned}
\end{equation}
and the final scalar reward is
\begin{equation}
r_{\mathrm{A1/A2}} = 0.10\,r_{\text{fmt}} + 0.35\,r_{\text{sem}} + 0.55\,r_{\text{num}}.
\end{equation}
Intuitively, the model only receives numeric credit when it first predicts the correct motion type and direction.

\paragraph{A3/D3: multi-step sequence rewards.}
For multi-step action prediction (A3) and object-guided action inference (D3), let ${\mathbf a}^\star=(a_1^\star,\dots,a_m^\star)$ be the ground-truth sequence and $\hat{\mathbf a}=(\hat{a}_1,\dots,\hat{a}_n)$ the parsed predicted sequence. The implementation defines
\begin{equation}
r_{\text{fmt}}=\mathbb{1}[n>0],
\end{equation}
\begin{equation}
r_{\text{sem}} = \frac{\sum_t \min(c_t(\hat{\mathbf a}), c_t({\mathbf a}^\star))}{\max(n,m,1)},
\end{equation}
where $c_t(\cdot)$ counts how many times action type $t$ appears in the sequence,
\begin{equation}
r_{\text{ord}} = \frac{1}{m}\sum_{i=1}^{\min(m,n)} \mathbb{1}[\hat{a}_i = a_i^\star],
\end{equation}
and
\begin{equation}
r_{\text{num}} = \frac{1}{m}\sum_{i=1}^{\min(m,n)} \mathbb{1}[\hat{a}_i = a_i^\star]\, s_{\mathrm{num}}(\hat{v}_i, v_i^\star; a_i^\star).
\end{equation}
The pre-penalty score is
\begin{equation}
\tilde{r}_{\mathrm{seq}} = 0.10\,r_{\text{fmt}} + 0.25\,r_{\text{sem}} + 0.35\,r_{\text{ord}} + 0.30\,r_{\text{num}},
\end{equation}
and the final sequence reward is
\begin{equation}
r_{\mathrm{A3/D3}} = \max\!\bigl(0,\; \tilde{r}_{\mathrm{seq}} - 0.03\max(0,n-m)\bigr).
\end{equation}
Thus the reward favors correct action content, correct order, and correct magnitudes, while mildly penalizing unnecessarily long predicted sequences.

\paragraph{A4/D2/D4: binary rewards.}
For action verification (A4), visibility judgment (D2), and object-identity verification (D4), the implementation uses a short-form boolean parser:
\texttt{yes/true} $\mapsto$ positive and \texttt{no/false} $\mapsto$ negative. The component rewards are
\begin{equation}
r_{\text{fmt}}=\mathbb{1}[\hat{y}\in\{\texttt{yes},\texttt{no},\texttt{true},\texttt{false}\}],
\quad
r_{\text{sem}}=\mathbb{1}[\hat{y}=y^\star],
\end{equation}
and the final reward is
\begin{equation}
r_{\mathrm{bin}} = 0.20\,r_{\text{fmt}} + 0.80\,r_{\text{sem}}.
\end{equation}
A practical limitation of this binary reward is that it only recognizes short-form labels. If the reference answer is expressed in an alternative form such as \texttt{A/B} or a full sentence, the reward no longer provides binary-accuracy credit. This limitation is acceptable for the current task setting, but it also suggests that a more format-robust binary reward would be desirable in future work.

\paragraph{D1: geometric box reward.}
For post-action object localization, the response is scored as a structured bounding box. The format term gives 1.0 if the response is exactly a bbox string of the form \texttt{[x1, y1, x2, y2]}, 0.4 if a parseable bbox appears anywhere in the response, and 0 otherwise. Let $\hat{\mathbf b}$ and ${\mathbf b}^\star$ denote the predicted and target boxes. The implementation first canonicalizes the coordinates by swapping endpoints when needed and clipping them to $[0,1000]$. It then defines
\begin{equation}
r_{\text{valid}} = 0.7\,\mathbb{1}[x_1 < x_2 \wedge y_1 < y_2] + 0.3\,\max\!\left(0, 1 - \frac{\mathrm{overflow}}{200}\right),
\end{equation}
where \textit{overflow} is the average out-of-range amount before clipping. After canonicalization, the localization-quality term combines four geometric measures:
\begin{equation}
r_{\mathrm{IoU}} = \mathrm{IoU}(\hat{\mathbf b}, {\mathbf b}^\star),
\end{equation}
\begin{equation}
r_{\text{center}} = \max\!\left(0, 1 - \frac{\lVert \hat{\mathbf c} - {\mathbf c}^\star \rVert_2}{\max(80,\; 0.6\,d^\star)}\right),
\end{equation}
where $d^\star$ is the diagonal length of the target box,
\begin{equation}
r_{\text{size}} = \max\!\left(0, 1 - \frac{|\log(\hat{w}/w^\star)| + |\log(\hat{h}/h^\star)|}{1.6}\right),
\end{equation}
and
\begin{equation}
r_{\text{L1}} = \max\!\left(0, 1 - \frac{1}{180}\cdot\frac{1}{4}\sum_{k=1}^4 |\hat{b}_k - b_k^\star|\right).
\end{equation}
These are aggregated as
\begin{equation}
r_{\text{geo}}^{\text{base}} = 0.45\,r_{\mathrm{IoU}} + 0.20\,r_{\text{center}} + 0.20\,r_{\text{L1}} + 0.15\,r_{\text{size}},
\end{equation}
\begin{equation}
r_{\text{geo}} = r_{\text{geo}}^{\text{base}} \cdot (0.3 + 0.7\,r_{\text{valid}}),
\end{equation}
and the final D1 reward is
\begin{equation}
r_{\mathrm{D1}} = 0.20\,r_{\text{fmt}} + 0.15\,r_{\text{valid}} + 0.65\,r_{\text{geo}}.
\end{equation}
Hence D1 is the only task where the geometric term is further decomposed into a validity sub-score and a localization-quality sub-score.

\paragraph{Task-aware reward structure.}
World2VLM does not use a single uniform symbolic reward across all tasks. Instead, each sample is routed to a task-specific parser, and only the terms relevant to that task are activated: numeric precision matters for A1/A2/A3/D3, order consistency only for sequence tasks, and geometric quality only for D1. This task-aware construction allows GRPO to refine structured outputs without forcing heterogeneous task types into a single reward template.

\section{Qualitative Analysis of GRPO Improvements}
\label{sec:appendix-grpo-analysis}

\paragraph{Qualitative role of GRPO.}
The aggregate gains of World2VLM-GRPO over World2VLM-SFT in Table~\ref{tab:main-results} establish the benefit of reinforcement learning at the benchmark level, but qualitative inspection is useful for characterizing \emph{how} the refinement stage changes model behavior. Table~\ref{tab:appendix-grpo-cases} therefore presents representative held-out examples spanning numeric regression, counting, discrete directional reasoning, and output regularization.

\begin{table*}[t]
\centering
\footnotesize
\setlength{\tabcolsep}{5pt}
\begin{tabular}{p{1.5cm}p{4.5cm}p{0.8cm}p{1.9cm}p{1.9cm}p{5.1cm}}
\toprule
\textbf{Benchmark} & \textbf{Question Type / Prompt} & \textbf{GT} & \textbf{SFT Output} & \textbf{GRPO Output} & \textbf{Observation} \\
\midrule
SPARBench & Depth prediction: estimate the depth of a cart given a box at 3.7\,m. & 5.4 & 10.5 & 5.1 & GRPO removes a severe overestimation and brings the prediction close to the ground-truth range. \\
SPARBench & Depth prediction: estimate the depth of toilet paper given a towel at 1.2\,m. & 1.8 & 0.6 & 1.5 & GRPO substantially reduces underestimation and yields a more plausible metric prediction. \\
VSI-Bench & Object counting: ``How many table(s) are in this room?'' & 2 & 1 & 2 & GRPO converts a near miss into the correct count. \\
VSI-Bench & Object-size estimation: estimate the longest dimension of a stool in centimeters. & 42 & 450 & 50 & GRPO corrects an order-of-magnitude error and produces a better calibrated numeric estimate. \\
SAT-Real & Perspective taking: ``If I stand where the soft toy is and face right by 90 degrees, is the cellphone to the left or right?'' & right & left & right & GRPO corrects the directional judgment in a representative real-image perspective-taking example. \\
MindCube & Option prediction from multi-view evidence. & C & \texttt{<answer>D. Decorated wall</answer>} & \texttt{C. Light brown wall} & GRPO improves both correctness and output regularity by removing the extra wrapper and aligning the option label with the correct semantic referent. \\
\bottomrule
\end{tabular}
\caption{Representative SFT$\rightarrow$GRPO output changes on held-out evaluation examples. GT denotes ground truth. The examples illustrate three recurring effects of GRPO in our setting: better numeric calibration, more consistent discrete decisions, and cleaner answer formatting.}
\label{tab:appendix-grpo-cases}
\end{table*}

\paragraph{Numeric precision and calibration.}
The clearest effect of GRPO appears on scalar regression or estimation tasks. In both SPARBench depth-prediction examples in Table~\ref{tab:appendix-grpo-cases}, the SFT model produces numerically legal answers, but the magnitudes are badly calibrated: one is a large overestimate (10.5 instead of 5.4), and one is a large underestimate (0.6 instead of 1.8). After GRPO, both predictions move markedly toward the correct range. A similar pattern appears in VSI-Bench size-estimation examples, where GRPO often suppresses order-of-magnitude failures such as predicting 450\,cm for an object whose ground-truth size is 42\,cm. This behavior is consistent with the explicit numeric reward terms described in Appendix~\ref{sec:appendix-grpo-reward}: once the model already knows the answer \emph{type}, GRPO appears to sharpen the magnitude itself.

\paragraph{Discrete decision consistency.}
GRPO also benefits tasks whose answers are discrete labels rather than free-form numbers. Qualitatively, the most visible improvements arise in categories such as egocentric movement, perspective taking, and object movement, where success depends on separating a small set of semantically similar alternatives. This pattern suggests that the GRPO stage mainly refines the final decision boundary once SFT has already provided a plausible latent spatial representation. By contrast, categories that require more demanding relational updates exhibit smaller qualitative changes, consistent with the view that GRPO primarily calibrates decision quality rather than replacing the transition knowledge learned during SFT.

\paragraph{Response regularization.}
Another recurring effect is improved alignment between model outputs and the task-specific answer schema. After GRPO, responses more often appear directly in the expected short form, with fewer extraneous wrappers or weakly normalized strings. This behavior is consistent with the format-sensitive reward terms and suggests that GRPO improves not only correctness, but also the evaluability and deployment robustness of the final response.

\paragraph{Interpretation.}
Taken together, these examples suggest that the role of GRPO in World2VLM is not to replace the transition knowledge already learned during SFT. Rather, its main contribution is to \emph{sharpen} that knowledge at the output layer: it improves numeric calibration, increases the reliability of discrete directional and multiple-choice decisions, and makes the final answer format more consistent with the expected task schema. This is exactly the division of labor we intended between the two stages: SFT internalizes motion-conditioned structure, and GRPO improves how that structure is expressed at inference time.

\section{Implementation Details}
\label{sec:appendix-implementation-details}

\paragraph{Supervised fine-tuning setup.}
The supervised stage performs parameter-efficient post-training of Qwen2.5-VL-7B-Instruct on 8 H100 GPUs using bfloat16, AdamW, cosine learning-rate decay, 5\% warmup, gradient accumulation 2, and ZeRO-2. We use seed 42, maximum sequence length 8192, and the native Qwen2.5-VL chat template for multimodal serialization.

\paragraph{Data split and visual preprocessing.}
SFT uses the mixed-source World2VLM training set of approximately 103K examples, with a 1\% held-out validation split for model selection. Images are processed by the standard Qwen2.5-VL visual processor with dynamic resizing and a bounded visual-token budget, preserving the model's native patch and merge settings while keeping multi-GPU training stable.

\paragraph{Adaptation scope.}
We adopt LoRA rather than full-parameter tuning. The visual tower remains frozen throughout SFT, and the adapters use rank 256, scaling 512, dropout 0.05, and no bias adaptation. The LoRA modules are applied to language-side attention and MLP projection layers, so the post-training focuses on multimodal reasoning behavior expressed through the language backbone rather than altering the vision encoder itself.

\paragraph{Evaluation protocol.}
Reported SFT evaluations use the merged checkpoint with near-greedy decoding, making generation effectively deterministic unless a benchmark wrapper specifies otherwise. Downstream evaluation is conducted with batch size 1 for LMMS-Eval benchmarks, and with the benchmark-specific image-budget settings used for SAT-family and MindCube evaluation.

\section{Baseline Implementation Details and Comparative Discussion}
\label{sec:appendix-baseline-details}

\paragraph{Baseline scope.}
The original MindJourney framework uses GPT-4o and other closed-source models in several roles within its inference-time imagination pipeline \cite{yang2025mindjourney}. For a fair comparison with World2VLM, we transfer the same test-time paradigm onto the same open Qwen2.5-VL backbone used by our method. The resulting baseline therefore matches the backbone and differs primarily in where world-model guidance is introduced: at inference time for the MindJourney-style baseline, and during post-training for World2VLM.

\paragraph{Experimental configuration.}
The baseline keeps the Qwen2.5-VL backbone frozen, uses a world model to render imagined views under candidate egocentric actions, asks the VLM to score those imagined views with respect to the original question, ranks candidate trajectories from those scores, and produces the final answer from the selected path. In the SAT-family setting analyzed here, search uses up to three imagination steps, beam size two, the canonical forward/left-turn/right-turn action families, and the same benchmark-side image budget used throughout our experiments. This configuration preserves the central MindJourney mechanism while maintaining a backbone-matched comparison.

\paragraph{Discussion.}
This matched-baseline design is informative because both methods start from the same Qwen2.5-VL model. The MindJourney-style baseline measures the benefit of invoking a world model during inference without updating the backbone, whereas World2VLM measures the benefit of distilling analogous spatial guidance into the model during training. Under this controlled comparison, the results suggest that training-time internalization offers a favorable accuracy-efficiency trade-off while preserving direct comparability to inference-time imagination.

\section{Teacher World-Model Quality and Failure Cases}
\label{sec:appendix-teacher-quality}

\paragraph{Importance of teacher quality.}
World2VLM treats the world model as a training-time teacher, so the quality of the generated supervision inevitably upper-bounds the quality of the distilled student. If the teacher produces images with visible artifacts, geometric distortion, action-inconsistent viewpoint changes, or unstable object identity across views, then the resulting QA supervision becomes noisy. In the mild case, this noise simply weakens the usefulness of a training sample; in the severe case, it can actively teach the student the wrong motion-to-observation mapping. This is especially important here because our supervision is not limited to natural-language captions: it also includes metric motion values, structured action sequences, object boxes, visibility judgments, and cross-view identity labels, all of which are sensitive to teacher errors.

\paragraph{Qualitative difference between SVC and HY-WorldPlay.}
Our experiments use two different teacher families precisely because no single world model is uniformly best along all dimensions. SVC is a camera-conditioned novel-view generator and tends to preserve explicit viewpoint-control semantics better, which is consistent with its stronger performance on SAT-Real and VSI-Bench in the main paper. However, in our generated trajectories SVC also more frequently exhibits local rendering artifacts under larger viewpoint changes, such as stretched edges, duplicated structures, warped object boundaries, and texture-level tearing around thin objects or cluttered depth discontinuities. HY-WorldPlay, by contrast, generally produces cleaner-looking frames and stronger perceptual realism, which qualitatively makes it a better image teacher in many cases. This stronger perceptual quality is consistent with its better downstream performance on SAT-Synthesized and MindCube. At the same time, HY-WorldPlay is not simply a uniformly better replacement: as an action-conditioned video world model, it can sometimes smooth, reinterpret, or visually regularize fine-grained camera motion in ways that make the rendered transition look plausible while being less tightly aligned with the exact camera-conditioned geometry that SVC was designed to model.

\paragraph{Propagation of teacher errors into supervision.}
Teacher errors affect different task families in different ways.
First, \textbf{action-observation mismatch} directly corrupts motion-centric tasks. If the rendered endpoint under-rotates, over-rotates, or drifts laterally relative to the nominal action trace, then A1, A2, A3, and A4 receive supervision whose target answer is formally correct with respect to metadata but visually inconsistent with the image pair.
Second, \textbf{geometric distortion} disproportionately harms object-grounded tasks. Warped boxes, broken object silhouettes, or cross-view size inconsistency can make D1 localization targets less meaningful and can also reduce the reliability of D3/D4, which depend on stable object correspondence across views.
Third, \textbf{appearance hallucination and disappearance errors} are particularly harmful for D2 and D4. If an object is spuriously removed, newly hallucinated, or semantically altered by the teacher, the student may be rewarded for a visibility or identity judgment that would be wrong in the underlying scene.
Fourth, \textbf{large-motion instability} is amplified by our max-displacement pairing strategy. This strategy is useful because it increases the supervision signal, but it also means that the paired source-target views are drawn from the part of the trajectory where teacher errors are often most visible.

\paragraph{Interpretation of the teacher comparison.}
The benchmark pattern in Table~\ref{tab:main-results} is consistent with the failure modes above. SVC appears most useful when exact camera-conditioned viewpoint consistency matters more than photorealistic detail, which helps explain its advantage on SAT-Real and VSI-Bench. HY-WorldPlay appears more useful when global perceptual plausibility, smoother appearance evolution, and stronger temporal priors help mental simulation, which is consistent with its stronger performance on SAT-Synthesized and MindCube. We therefore do not interpret the teacher comparison as ``one teacher is universally better.'' Instead, the results suggest that different teacher families inject different spatial biases into the student: SVC contributes tighter camera-geometry alignment, while HY-WorldPlay contributes stronger perceptual realism and action-conditioned temporal continuity.

\paragraph{Current mitigation strategies.}
Although we do not yet perform explicit teacher-reliability modeling, several parts of the current pipeline already act as partial safeguards against teacher noise. We use two different teacher families rather than depending on a single generator, which reduces the chance that the full training signal inherits one model's idiosyncratic errors. We also filter object-level supervision using confidence, area-ratio, border-margin, and track-consistency constraints, so badly degraded teacher outputs are less likely to survive into D1--D4. In addition, prompt-level validation removes malformed records, and mixed-source training across real and simulated anchors reduces overfitting to one teacher-domain combination. Finally, the GRPO stage encourages short, well-formed, task-structured outputs, which helps the student avoid compounding noisy supervision with verbose but weakly grounded responses.

\paragraph{Limitations and future directions.}
Even with these safeguards, teacher quality remains a real bottleneck. World2VLM currently treats all retained world-model-generated samples as equally trustworthy once they pass rule-based filtering. A stronger version of the framework should estimate teacher reliability more explicitly, for example by down-weighting samples with teacher-student disagreement, filtering trajectories with weak cross-view geometric consistency, using detector confidence as a soft reliability signal rather than a hard gate, or requiring agreement between multiple teacher rollouts before accepting a sample. Another promising direction is to make teacher quality task-aware: exact camera-motion tasks may prefer a camera-consistent teacher such as SVC, whereas mental-simulation tasks may benefit more from a perceptually stronger temporal teacher such as HY-WorldPlay. We view such quality-aware teacher selection and confidence-weighted distillation as an important next step for improving the robustness of training-time world-model alignment.

\end{document}